\documentclass[10pt,journal,twoside]{IEEEtran}

\usepackage[utf8]{inputenc}
\usepackage[T1]{fontenc}
\usepackage[english]{babel}
\usepackage[linesnumbered]{algorithm2e}
\usepackage{amsmath,amsfonts,amssymb,amsthm,stix}
\usepackage{mathtools}
\usepackage{acronym}
\usepackage{textcomp}
\usepackage{multirow}
\usepackage{makecell}
\usepackage[caption=false,font=footnotesize]{subfig}
\usepackage{graphicx}
\usepackage{url}
\usepackage{balance}
\usepackage{float}
\usepackage{pifont}
\usepackage{array}
\usepackage{bm}
\usepackage{booktabs}
\usepackage{enumitem}
\usepackage{tikz}
\usepackage{xcolor}
\usepackage[hidelinks]{hyperref}

\graphicspath{{figures/}}

\newcommand*\circled[1]{\tikz[baseline=(char.base)]{
            \node[shape=circle,fill=black,draw,inner sep=1pt] (char) {#1};}}

\newcommand{\sol}
{VESSI}

\SetCommentSty{mycommfont}
\newcolumntype{P}[1]{>{\centering\arraybackslash}m{#1}}
\acrodef{VLM}{Vision Language Model}
\acrodef{LLM}{Large Language Model}
\acrodef{CTI}{Cyber Threat Intelligence}
\acrodef{CAGR}{Compound Annual Growth Rate}
\acrodef{CMUS}{Composite Model Utility Score}
\acrodef{SOC}{Security Operations Center}
\acrodef{CUDA}{Compute Unified Device Architecture}
\acrodef{AI}{Artificial Intelligence}
\acrodef{GPU}{Graphics Processing Unit}
\acrodef{SOTA}{State-of-the-Art}
\acrodef{OCR}{Optical Character Recognition}
\acrodef{UCF}{University of Central Florida}
\acrodef{DRAM}{Dynamic Random Access Memory}
\acrodef{VRAM}{Video Random Access Memory}
\acrodef{NLP}{Natural Language Processing}
\acrodef{API}{Application Programming Interface}
\acrodef{CSV}{Comma Separated Values}
\acrodef{UAV}{Unmanned Aerial Vehicle}
\acrodef{NVML}{NVIDIA Management Library}
\acrodef{FPS}{Frames Per Second}
\acrodef{MLP}{Multilayer Perceptron}
\acrodef{VIRAT}{Video and Image Retrieval and Analysis Tool}
\acrodef{UCA}{Understanding Crime Activities}
\acrodef{TVQA}{TV Question Answering}
\acrodef{VLPD}{Vision-Language Pedestrian Detection}
\acrodef{CPU}{Central Processing Unit}

\begin{document}

\title{VESSI - VLM-Enhanced Support for Surveillance and Investigations}

\author{Saverio Cavasin, Pietro Tedeschi, Mattia Tamiazzo, Alessandro Brighente,
        Simone Milani, and Mauro Conti%
\thanks{S. Cavasin, A. Brighente, and M. Conti are with the Department of Mathematics, University of Padua, Padua, Italy.}%
\thanks{S. Cavasin and P. Tedeschi are with the Innovation \& Research Center, CY4GATE S.p.A., part of ELT Group, Rome, Italy.}%
\thanks{M. Tamiazzo and S. Milani are with the Department of Information Engineering, University of Padua, Padua, Italy.}%
\thanks{M. Conti is also with \"{O}rebro University, \"{O}rebro, Sweden.}%
\thanks{Corresponding author: Saverio Cavasin (e-mail: saverio.cavasin@phd.unipd.it). E-mail addresses: pietro.tedeschi@cy4gate.com (P. Tedeschi), alessandro.brighente@unipd.it (A. Brighente), simone.milani@dei.unipd.it (S. Milani), and mauro.conti@unipd.it (M. Conti).}}

\markboth{IEEE Transactions on Information Forensics and Security}%
{Cavasin \MakeLowercase{\textit{et al.}}: VESSI - VLM-Enhanced Support for Surveillance and Investigations}

\maketitle

\begin{abstract}
Automated video surveillance analysis has become a critical component of intelligence infrastructures and Law Enforcement agencies. Traditional systems lack the semantic module for comprehensive situational awareness and forensic tasks, limiting their ability to interpret events meaningfully or support post-incident investigations. This slows operational insight and increases the burden on human analysts.
Recent advances in the Vision-Language Model (VLM) offer promising pathways to bridge this gap.
To address this, we propose {VLM-Enhanced Support for Surveillance and Investigations (VESSI)}, a VLM-based framework designed to enhance automated video surveillance analysis through prompt-driven interrogation of video sequences where salient visual features are converted into textual descriptions.

We test our framework with four state-of-the-art models.
Since most of the datasets for this task are unlabeled, we also propose the metric \ac{CMUS} to assess VLMs' performance.
{Experimental results show that our solution substantially improves the analysis capabilities of human operators and enhances the flexibility of automated surveillance systems. In our evaluation, the most reliable model flagged potentially relevant activity in more than $66\%$ of the videos while reducing review time by more than $85\%$, offering a practical balance between selectivity and efficiency. The model ordering produced by the reference-free \ac{CMUS} evaluation was reproduced by the normal-video \ac{CMUS} evaluation and matched the false-positive-rate ordering obtained from $5{,}909$ manually referenced frames. This agreement supports the operational use of the score within the evaluated setting.}
\end{abstract}


\section{Introduction}
\label{sec:intro}
{
The growth of urban video surveillance, projected to reach a market value of USD 81.37 billion by 2030~\cite{gvr2025_cameras}, is producing an increasing volume of footage for public safety, crime prevention, traffic monitoring, and infrastructure management. Forensic use of this material still relies heavily on manual inspection, which is slow, resource-intensive, and vulnerable to error when large collections must be reviewed under time pressure.

\acp{VLM} can query visual content through natural language and support scene summarization~\cite{shao_2018_eccv}, event interpretation, and interactive search~\cite{feng_2020_wacv,wang_2025_iclr}. These capabilities can reduce the cognitive and temporal burden on human operators~\cite{seidel_2025_drones} and make surveillance analysis more scalable, inspectable, and semantically expressive. Their use in operational systems, however, raises two main issues. First, current models require substantial memory and computational resources~\cite{xu_2024_cvprws,nassef_2024_jes}, especially on embedded devices or distributed camera networks. Compression~\cite{zhu_2024_tacl}, cloud offloading~\cite{ullah2023_jcc}, and hardware acceleration~\cite{Burhanuddin2023_kharizmia} can mitigate this cost, with different effects on latency and scalability. Second, \acp{VLM} can produce inconsistent or hallucinated outputs on ambiguous and out-of-distribution inputs~\cite{rawte_2025_trustnlp,zollicoffer_2025_neurips}. Such failures are consequential in forensic settings~\cite{dahl_2025_judicature}, while continuous manual validation can erode the expected efficiency gain~\cite{gunjal_2024_aaai}.

Existing large-model approaches to video understanding often retain a narrow focus on object detection and tracking~\cite{tang_2024_acl}. Structured querying, traceable frame retrieval, operational profiling, and evaluation without exhaustive frame-level labels remain insufficiently addressed. This work targets this gap through a modular surveillance-triage pipeline and an evaluation method designed for ambiguous, video-level-labeled material.
}

\subsection{Contributions}
{This work proposes a \ac{VLM}-based surveillance framework that combines semantic querying, operational feasibility, and traceable outputs. It also examines deployment constraints and comparative reliability in sensitive settings such as law enforcement, drone monitoring, and defense.}

In a nutshell, our main contributions can be summarized as follows.

\begin{itemize}
    \item We propose a lightweight modular architecture for single-frame surveillance interrogation, semantic prompting, and traceable evidence retrieval with \acp{VLM}.

    \item We introduce the \textbf{Composite Model Utility Score} (CMUS), which combines detection behavior, inference efficiency, and hallucination control when exhaustive frame-level ground truth is unavailable.

    \item We define technical guidelines for adapting the framework to cloud-based \ac{SOC} infrastructures and constrained edge systems.

    \item We benchmark four 7B-parameter \acp{VLM}, identifying trade-offs among semantic selectivity, latency, hardware demand, and robustness.
\end{itemize}

It is possible to outline the research questions answered by this paper in a four-point list.

\begin{itemize}[leftmargin=11mm]
    \item[\textbf{(RQ1)}] Can \acp{VLM} effectively improve Law Enforcement frameworks in video surveillance analysis?
    \item[\textbf{(RQ2)}] Which \ac{SOTA} architecture and model is more suitable to support Law Enforcement in surveillance?
    \item[\textbf{(RQ3)}] Which information and strategies should be implemented to consistently pair \ac{VLM} agents in surveillance?
    \item[\textbf{(RQ4)}] Can we effectively test \ac{VLM} models even when surveillance material is not frame-by-frame labeled (usual case for data of this nature)?
\end{itemize}

\subsection{Roadmap}
The paper is organized as follows.
Section~\ref{sec:related} reviews the literature on vision-language surveillance and positions \sol\ with respect to existing systems.
Section~\ref{sec:models} introduces the four models considered in the study.
Section~\ref{sec:method} presents the \sol\ pipeline and its acquisition, analysis, and threat-evaluation phases.
Section~\ref{sec:experiments} describes the experimental protocol, while Section~\ref{sec:cmus} defines the reference-free evaluation measures and the \ac{CMUS} score.
Section~\ref{sec:results} reports detection, validation, control-group, utility, and computational results.
{Section~\ref{sec:future} discusses the limitations and future directions, and Section~\ref{sec:conclusion} concludes the paper.}
The notation used throughout the paper is summarized in the supplementary material.

\section{Related Work and Comparison}
\label{sec:related}
The integration of large \aclp{VLM} into surveillance video analysis is an emerging and rapidly evolving research area.
Although many recent efforts focus on general-purpose video understanding, the application of these systems to high-stakes, real-world surveillance tasks, especially with forensic or cybersecurity constraints, remains limited and underexplored.

\subsection{Related works}

Sultani \textit{et al.}~\cite{sultani2018_cvpr} introduced the \ac{UCF}-Crime dataset, one of the first large-scale benchmarks aimed at anomaly detection in untrimmed surveillance videos.
Their approach focused on weakly supervised learning through multiple-instance ranking to localize anomalies using only video-level labels.
This dataset has become a standard reference in anomaly detection research, but does not include natural language annotations or interactive querying for in-depth analysis.

Building on \ac{UCF}-Crime, Yuan \textit{et al.}~\cite{yuan2024cvpr} proposed the \ac{UCA} benchmark, designed to extend surveillance video understanding with temporally grounded language annotations.
Their framework supports video-language tasks such as temporal sentence grounding, event captioning, and question answering.
However, it requires dense spatiotemporal modeling and purpose-trained models such as surveillance \textit{SwinBERT}~\cite{lin2022swinbert}.
This limits modularity and flexibility, especially in contexts requiring resource-awareness or {output traceability}, and does not offer different model comparisons.

\ac{TVQA}~\cite{lei_2018_emnlp} further advanced video-language QA approaches by localizing answers to temporal segments.
Nevertheless, their study focuses on a dataset built using television series, movies, and scripted videos.
The lower intrinsic quality of surveillance videos, inherited in their nature, makes this study insufficient for real-world applications.

The \ac{VIRAT} dataset~\cite{oh2011_cvpr} is one of the most naturalistic collections of real-world surveillance data, emphasizing long, continuous scenes with complex human and vehicle activities.
However, it lacks any multimodal or language interface, rendering it unsuitable for direct evaluation of \ac{VLM}-based semantic reasoning or forensic tasks.

Other \ac{VLM}-powered approaches have also emerged in adjacent domains. For example, \textit{Clip2Safety}~\cite{chen2025_eswa} and \ac{VLPD}~\cite{liu_2023_cvpr} target workplace safety and pedestrian detection, respectively, using visual grounding at the scene level. Meanwhile, applications like \textit{LlavaGuard}~\cite{helff_2025_icml} focus on image-text moderation. Although demonstrating the spread interest and need for such technologies, these systems are not tailored for forensic investigation or flexible integration into lightweight, field-deployable environments.

Regarding the analysis of real-world data without frame-level ground truth, a similar approach leveraging structured prompt templates is presented in VIVID~\cite{gonzalez_2025_visapp}.
It supports the semantic-guided frame interrogation in surveillance footage, but it is only applied to identify generic violence by skipping other criminal activities, making it suitable to categorize salient or specific events.

{
Recent studies have connected vision-language reasoning more directly to video anomaly detection.
LAVAD~\cite{zanella2024lavad} follows a training-free pipeline in which frame captions are temporally aggregated and converted into anomaly scores by a large language model.
VERA~\cite{ye2025vera} learns verbalized guiding questions from coarsely labeled videos and uses them with a frozen \ac{VLM} to produce interpretable anomaly scores.
ASK-HINT~\cite{zou2026askhint} studies structured, action-centric prompting and shows that the information and granularity encoded in the prompt affect anomaly-detection performance.
CPVAD~\cite{deng2026cpvad} first extracts context-preserving temporal boundaries and then performs clip-level anomaly scoring with a frozen \ac{VLM}; on XD-Violence, its boundary-based sampling reduces the reported number of model queries from $146{,}449$ to $5{,}340$ while retaining temporal context.
These methods primarily address anomaly scoring and temporal localization over clips.
\sol\ addresses a complementary operational setting based on category-specific frame interrogation, controlled prompt-complexity analysis, multi-model profiling, and traceable storage of flagged frames, timestamps, and model responses.
}

{
VANE-Bench~\cite{gani2025vanebench} provides a dedicated benchmark for evaluating video-language models through anomaly-detection and localization questions across synthetic and real-world videos, including crime-related footage.
Its evaluation of nine video-language models offers useful evidence on the capabilities and limits of general-purpose video question answering for anomaly understanding, while its scope remains benchmark-oriented and does not define an operational forensic-surveillance pipeline.
}

Concerning the effectiveness of \acp{VLM},  VALOR-EVAL~\cite{qiu_2024_valoreval} provides a comprehensive benchmark for vision-language models that balances multi-dimensional performance metrics within a unified evaluation framework.

However, while VALOR-EVAL also aims to balance multiple performance aspects, specifically faithfulness and coverage in open-vocabulary captioning, it is designed for static image evaluation and assumes fine-grained, human-annotated ground truth. By contrast, \acl{CMUS} is explicitly tailored to surveillance video analysis, where such ground truth is unavailable, and evaluation must account for detection rate and activation patterns over time.

Moreover, our hallucination scoring penalizes persistent over-activation. It does not evaluate semantic mismatches in captions, making it more suitable for forensic use cases where excessive alerts carry operational costs. \acl{CMUS} thus fills a distinct gap by enabling practical, lightweight performance assessment in low-supervision, streaming contexts such as long-term video surveillance.

\subsection{Comparison with the State of the art}

To the best of our knowledge, \sol\ is the only solution fulfilling, at the same time, all the considered requirements. Moreover, the architecture of \sol\ allows independent frame querying through structured prompts, providing fine-grained control, inspectable output records, and adaptability across hardware-constrained environments such as drones or embedded cameras. Our framework is explicitly designed to acknowledge the operational risks and failure points of modern \acp{VLM}, including hallucinations and adversarial susceptibility~\cite{wickramasekara_2025_fsid}. Finally, by addressing the gaps in existing datasets and \ac{VLM} deployments, our system supports flexible, lightweight, and traceable scene interrogation for real-world surveillance video, with direct relevance to digital forensics and cyber-physical threat response. Table~\ref{tab:sota_comparison} summarizes the comparison with the previous \ac{SOTA}. Since the other works tackled the problem with different goals (e.g., scene description or generic anomaly detection) we cannot provide a direct numeric comparison of results. Nevertheless, we highlight the main differences and improvements we bring with respect to the previous works. {The symbols report capabilities explicitly implemented or experimentally assessed in each work; architectural compatibility alone is not treated as full fulfillment.}

\begin{table*}[t]

\caption{{Comparison of \sol\ with surveillance datasets, benchmarks, and recent \ac{VLM}-based anomaly-detection systems.} \\ \textit{Legend:} $\mdlgblkcircle$~Feature fully fulfilled, $\circlelefthalfblack$~Feature partially fulfilled, $\mdlgwhtcircle$~Feature not fulfilled.}
\label{tab:sota_comparison}
\centering
\scriptsize
\resizebox{\textwidth}{!}{%
\begin{tabular}{|p{2.45cm}|P{1.3cm}|P{1.3cm}|P{1.3cm}|P{1.3cm}|P{1.3cm}|P{1.3cm}|P{1.3cm}|P{1.3cm}|P{1.3cm}|}
\hline
\textbf{Feature / Capability} & \textbf{UCF-Crime}~\cite{sultani2018_cvpr} & \textbf{UCA}~\cite{yuan2024cvpr} & \textbf{VIRAT}~\cite{oh2011_cvpr} & {\textbf{LAVAD}~\cite{zanella2024lavad}} & {\textbf{VERA}~\cite{ye2025vera}} & {\textbf{VANE-Bench}~\cite{gani2025vanebench}} & {\textbf{ASK-HINT}~\cite{zou2026askhint}} & {\textbf{CPVAD}~\cite{deng2026cpvad}} & \textbf{\sol} \\
\hline
Multimodal Data Analysis (video + language) & $\mdlgwhtcircle$ & $\mdlgblkcircle$ & $\mdlgwhtcircle$ & {$\mdlgblkcircle$} & {$\mdlgblkcircle$} & {$\mdlgblkcircle$} & {$\mdlgblkcircle$} & {$\mdlgblkcircle$} & $\mdlgblkcircle$ \\
\hline
Frame-level Semantic Analysis & $\mdlgwhtcircle$ & $\mdlgblkcircle$ & $\mdlgwhtcircle$ & {$\mdlgblkcircle$} & {$\mdlgblkcircle$} & {$\circlelefthalfblack$} & {$\circlelefthalfblack$} & {$\circlelefthalfblack$} & $\mdlgblkcircle$ \\
\hline
Interactive prompt-based querying & $\mdlgwhtcircle$ & $\mdlgwhtcircle$ & $\mdlgwhtcircle$ & {$\mdlgwhtcircle$} & {$\mdlgwhtcircle$} & {$\circlelefthalfblack$} & {$\mdlgwhtcircle$} & {$\mdlgwhtcircle$} & $\mdlgblkcircle$ \\
\hline
Real-time forensic applicability & $\mdlgblkcircle$ & $\mdlgwhtcircle$ & $\mdlgwhtcircle$ & {$\circlelefthalfblack$} & {$\mdlgwhtcircle$} & {$\mdlgwhtcircle$} & {$\mdlgwhtcircle$} & {$\circlelefthalfblack$} & $\mdlgblkcircle$ \\
\hline
Lightweight \ac{SOTA} model deployed ($\leq$ 7B) & $\mdlgwhtcircle$ & $\mdlgwhtcircle$ & $\mdlgwhtcircle$ & {$\mdlgwhtcircle$} & {$\mdlgwhtcircle$} & {$\mdlgwhtcircle$} & {$\mdlgblkcircle$} & {$\mdlgwhtcircle$} & $\mdlgblkcircle$ \\
\hline
No manual annotation requirement & $\mdlgblkcircle$ & $\mdlgwhtcircle$ & $\mdlgblkcircle$ & {$\mdlgblkcircle$} & {$\circlelefthalfblack$} & {$\mdlgwhtcircle$} & {$\mdlgblkcircle$} & {$\mdlgblkcircle$} & $\mdlgblkcircle$ \\
\hline
Multi-model \ac{SOTA} Benchmarking & $\mdlgwhtcircle$ & $\mdlgwhtcircle$ & $\mdlgwhtcircle$ & {$\mdlgwhtcircle$} & {$\circlelefthalfblack$} & {$\mdlgblkcircle$} & {$\mdlgblkcircle$} & {$\circlelefthalfblack$} & $\mdlgblkcircle$ \\
\hline
Prompt Complexity Evaluation & $\mdlgwhtcircle$ & $\mdlgwhtcircle$ & $\mdlgwhtcircle$ & {$\mdlgwhtcircle$} & {$\circlelefthalfblack$} & {$\mdlgwhtcircle$} & {$\mdlgblkcircle$} & {$\mdlgblkcircle$} & $\mdlgblkcircle$ \\
\hline
\end{tabular}
}
\end{table*}

\section{Tested \aclp{VLM}}
\label{sec:models}
A Vision Language Model (VLM) is a type of machine learning architecture that is designed to understand and generate both visual and textual information.
Unlike a language model (which processes only text information), VLMs combine images and text data in a multimodal way to interpret visual content (e.g. photos or videos) or generate images from textual prompts.
To evaluate the applicability of \acp{VLM} in surveillance contexts, we selected four novel and high-performing models, each with approximately 7 billion parameters.
This setup enables a balanced benchmark across models with similar computational requirements, improving fair comparisons in terms of performance, {output consistency}, and modularity.
The selected models are the following:

\begin{itemize}
    \item \textbf{LLaVA-1.5}~\cite{liu2023improvedllava}: \textit{Large Language and Vision Assistant} (LLaVA) 1.5 integrates a \textit{CLIP-based} vision encoder, namely ViT-L/14~\cite{radford2021clip} with a Vicuna~\cite{chiang2023vicuna} language backbone, connected via a lightweight Multilayer Perceptron projection module.
    \textit{LLaVA-1.5} enhances this architecture (i.e., combining the vision encoder with the language model using a \acfp{MLP} projector) with improved alignment and instruction tuning, allowing multimodal dialogue capabilities and visual grounding.

    \item \textbf{InstructBLIP}~\cite{dai_2023_neurips}: \textit{InstructBLIP} is built on the BLIP-2 framework by injecting instruction-following capabilities through additional supervised fine-tuning.
    It uses a frozen vision encoder, i.e., \textit{ViT-G/14}, and a \textit{Query Transformer} (Q-Former) module that connects visual tokens to a pre-trained language model such as \textit{FLAN-T5}~\cite{chung2024flan_scaling} or Vicuna.

    \item \textbf{Qwen2.5-VL-7B-Instruct}~\cite{wang_2024_Qwen2V_arxiv}: \textit{Qwen2.5-VL} is a bilingual vision-language model developed by Alibaba~\cite{alibaba2025qwen2.5vl}, supporting both Chinese and English.
    The \textit{VL-7B-Instruct} variant is instruction-tuned and capable of multi-turn interactions and fine-grained image understanding \cite{huang_2023_neurips}.
    It employs a Transformer-based visual encoder combined with a large language decoder, optimized for open-ended multimodal tasks.

    \item \textbf{DeepSeek-VL-7B-Chat}~\cite{lu2024deepseekvl}: \textit{DeepSeek-VL} is an open-source \ac{VLM} designed for real-world vision-language understanding.
    The \textit{7B-Chat} variant utilizes a hybrid vision encoder combining \textit{SigLIP-L}~\cite{tschannen2025siglip2} and \textit{SAM-B}~\cite{kirillov2023segmentanything}, supporting high-resolution images up to $1024\times1024$ pixels. It is constructed based on the \textit{DeepSeek-LLM-7B} base model, trained on approximately $2$ trillion text tokens and $400$ billion vision-language tokens.
\end{itemize}

The decision to focus on models with a similar number of parameters derives from both technical and operational considerations.
These architectures and model sizes offer a good balance between model capability and computational feasibility with respect to their deployment in \acp{SOC}~\cite{liu_2022_soc_ai}, cloud-based surveillance systems, or high-end embedded environments.
Their popularity in public repositories also reflects a degree of community validation and stability in terms of usability and support.
We selected frame-by-frame image-based \acp{VLM}. Models designed for continuous video input, including Video-LLaVA~\cite{lin_2024_emnlp}, were excluded from the benchmark.
This decision is made for several reasons:

\begin{itemize}
    \item \textbf{Flexibility.} Frame-level analysis allows a flexible integration with other independent video pre-processing stages~\cite{wen_2024_cvpr} (e.g., shot segmentation, motion detection, scene selection), making the pipeline more robust and independent (indeed, video pre-processing may vary depending on the recording platform thus affecting the training procedure with bias).
    Furthermore, prompting individual frames provides a fine-grained control over the type of reasoning applied, like scene description vs. behavioral pattern analysis, which is harder to achieve with end-to-end video-language transformers.

    \item \textbf{Efficiency.} Analyzing frame-level output improves computational efficiency, reduces processing latency, and avoids potential disalignment~\cite{wang_2025_iclr} across different model configurations.
\end{itemize}

At the time of writing, many video-level \acp{VLM} are either not publicly released, require significantly more computational effort~\cite{bertasius_2024_cvprws}, or present limitations in terms of modularity~\cite{min_2024_morevqa}, making it harder to achieve a fair \ac{SOTA} comparison for benchmarking performances.
Although this frame-based approach may not fully capture temporal continuity, it provides a practical and extensible foundation for testing \ac{VLM} robustness in surveillance scenarios.

\section{The \sol\ Framework}
\label{sec:method}
Our framework, namely \sol, is designed as a modular and extensible pipeline for analyzing surveillance footage using \ac{VLM}.

\begin{figure}[htp]
    \centering
    \includegraphics[width=1\columnwidth , trim={1cm 2cm 0.8cm 1.5cm},clip]{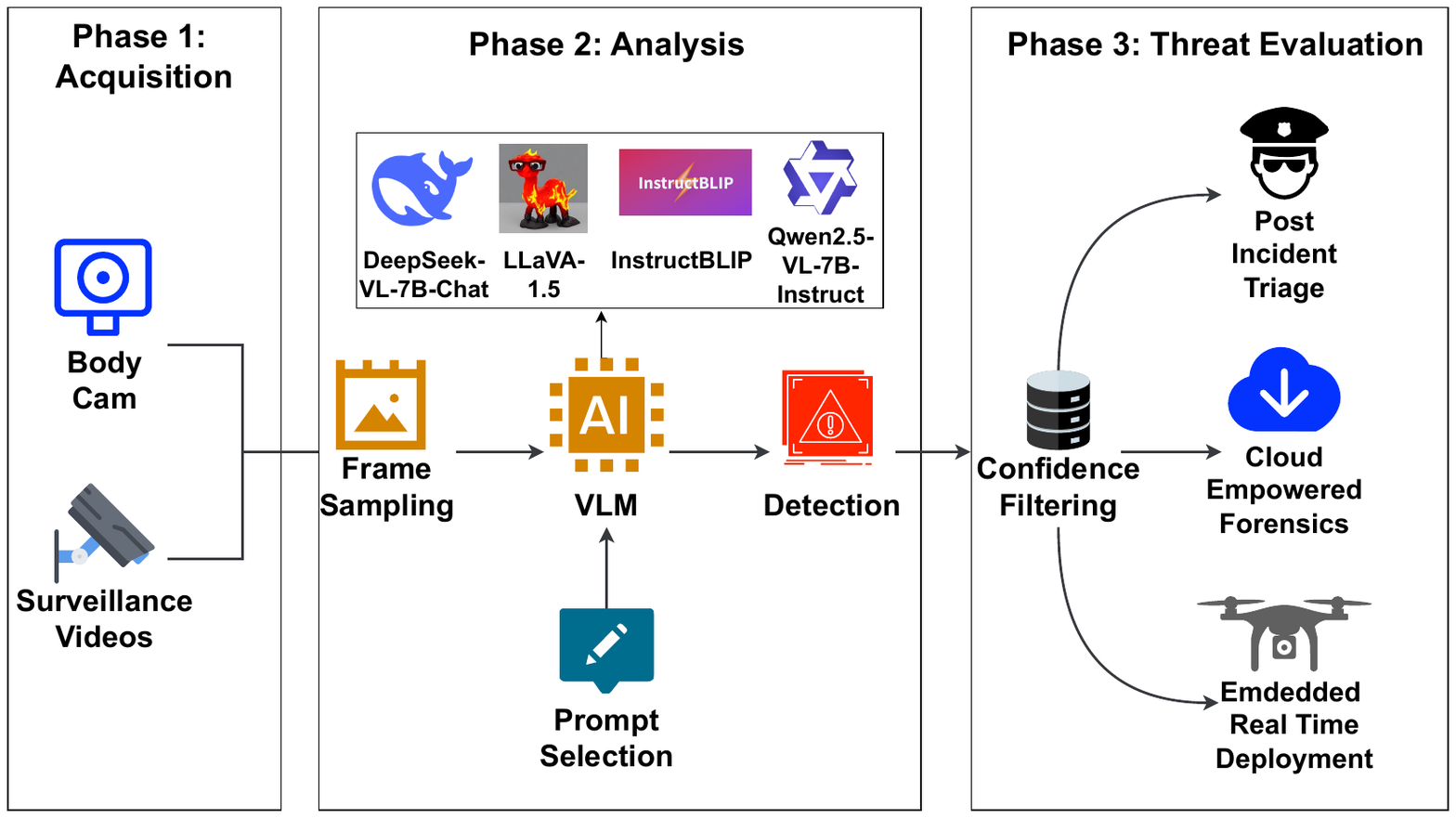}
    \caption{Framework Diagram of \sol.}
    \label{fig:diagram}
\end{figure}

Figure~\ref{fig:diagram} depicts the block diagram of our system, which can be  structured over three phases: the \emph{Acquisition}, the \emph{Analysis}, and the \emph{Threat Evaluation}, described in the following.

\subsection{Scope and Assumptions}
\label{sec:scope_assumptions}
{
The \ac{UCF}-Crime dataset provides video-level category labels but no exhaustive frame-level ground truth.
Individual frames can be ambiguous, depict overlapping activities, omit the labeled event because it occurs outside the camera view, or contain acquisition artifacts.
Accordingly, \sol\ is evaluated as a surveillance-triage framework: its scores describe operational utility and relative model behavior under these assumptions. They do not provide exhaustive temporal localization or objective frame-level accuracy.
}

\subsection{Phase 1 - Acquisition}
The input to the system is a categorized video dataset $\mathcal{V} = \{\mathcal{V}_1, \dots, \mathcal{V}_{13}\}$, where each subset $\mathcal{V}_c$ contains all videos $v$ belonging to the $c$-th crime class.

To properly assess the capabilities and performance of \sol, we adopt the \ac{UCF}-Crime dataset~\cite{sultani2018_cvpr}, a large-scale surveillance video collection organized into \textit{real-world crime}-related classes, as detailed in Table~\ref{tab:dataset_classes}.

\begin{table}[htpb]
\centering
\caption{Dataset classes, number of videos, and total duration (in minutes) per class.}

\label{tab:dataset_classes}
\begin{tabular}{lcc}
\toprule
\textbf{Class} & \textbf{Number of Videos} & \textbf{Total Duration (min)} \\
\midrule
Abuse           & $50$  & $107.50$  \\
Arrest          & $50$  & $165.22$  \\
Arson           & $50$  & $151.06$  \\
Assault         & $50$  & $72.19$  \\
Burglary        & $100$  & $261.78$  \\
Explosion       & $50$  & $140.22$  \\
Fighting        & $50$  & $143.83$  \\
Road Accidents  & $150$ & $144.93$  \\
Robbery         & $150$ & $234.77$  \\
Shooting        & $50$  & $81.93$   \\
Shoplifting     & $50$  & $180.20$  \\
Stealing        & $100$ & $259.68$  \\
Vandalism       & $50$  & $81.76$   \\
\midrule
Total           & $950$ & $2025.07$ \\
\bottomrule
\end{tabular}
\end{table}

It is worth noticing that the dataset is not manually annotated for training or fine-tuning.
We qualitatively assess the model outputs based on the relevance of the answers generated for the label class and the visual content of the frames selected by the algorithm given the textual query.

\subsection{Phase 2 - Analysis}
In this phase, the system processes each video $v \in \mathcal{V}_i$ by extracting a frame set $\mathcal{F}$ using uniform sampling at a fixed rate $r = 1$ Hz.
The goal is to evaluate how well different \acp{VLM} detect relevant visual content, given category-specific natural language prompts $p \in \mathcal{P} = \{p_1, p_2, p_3\}$.

\begin{enumerate}[label=\protect\circled{\color{white}\arabic*}]
\item Videos from each class are preprocessed by uniformly sampling frames at fixed intervals, with a tunable frame rate that we set to $1$ Hz for the initial evaluation.

\item Each frame $f \in \mathcal{F}$ is evaluated using a category-specific natural language prompt $p$, designed to test the model's ability to recognize semantically salient content. For each frame, we apply three prompts $p_1, p_2, p_3 \in \mathcal{P}$ of increasing semantic complexity, generated from the video-level category label:

\begin{itemize}[align=left]
    \item[\textbf{Easy.}] Direct binary query (e.g., \textit{``Do you see an act of \textbf{fighting} in the frame?''})
    \item[\textbf{Medium.}] Broader, less specific phrasing (e.g., \textit{``Does the frame show something that could be interpreted as an instance of \textbf{fighting}?''})
    \item[\textbf{Hard.}] Indirect, reasoning-intensive question (e.g., \textit{``Could a human observer reasonably conclude that this frame depicts \textbf{fighting}-like behavior or consequences?''})
\end{itemize}

Each query is deliberately formulated as a short binary decision. Its purpose is to provide a rapid and direct first-stage detection while limiting output-token generation and response latency. The constrained output also enables deterministic parsing and produces comparable decisions across models, prompts, and frames. \textit{Yes} and \textit{No} are the two requested decisions; responses such as ``uncertain,'' or any output that cannot be parsed as either decision, are treated as abstentions and normalized as \textit{Unclear}. An \textit{Unclear} response does not trigger an alert. A positive flag is also not a determination that a crime occurred: it selects the corresponding frame for subsequent investigative assessment by a human operator, such as a law-enforcement officer.

Each frame-prompt pair $(f, p)$ is passed through the model $\mathscr{m} \in \mathcal{M}$ in inference mode, where $\mathcal{M}$ is the set of the four tested models. At the same time, metrics such as inference time, memory usage, and sampled \ac{GPU} power draw are logged and exported for later analysis. This procedure is repeated across all selected models and prompt levels to ensure consistency in experimental comparisons.

{The three prompt formulations preserve the same binary detection task while varying the semantic complexity of the question. This design allows us to examine whether model outputs remain stable when the same investigative need is expressed through direct, broader, or reasoning-intensive language. }

\begin{algorithm}[!t]
\caption{Prompt-Based \ac{VLM} Frame Evaluation Pipeline.}
\label{alg:vlm_pipeline}
\SetAlgoLined
\KwIn{Video $v \in$ categorized video dataset $\mathcal{V} = \{ \mathcal{V}_1, \dots, \mathcal{V}_{13} \}$; prompt set $\mathcal{P} = \{p_1, p_2, p_3\}$; model $\mathscr{m} \in \mathcal{M}$; sampling rate $r$; confidence threshold $\tau$}
\KwOut{Per-frame response log $\mathcal{L}$ and salience-filtered alert frames}

\tcp{\small Phase 1 - Acquisition.}
$\mathcal{L} \gets \emptyset$ \tcp*[l]{\small Initialize Log}
\ForEach{category $c = 1$ \KwTo $13$}{
    \ForEach{video $v \in \mathcal{V}_c$}{
        $\mathcal{F} \gets \texttt{Extract\_Frame\_Set}(v, r)$\;

        \tcp{\small Phase 2 - Analysis.}
        \ForEach{frame $f \in \mathcal{F}$}{
            \ForEach{prompt $p \in \mathcal{P}$}{
                $o \gets \mathscr{m}(f, p)$\tcp*[l]{\small Model Inference}
                $(\text{label}, \zeta) \gets \texttt{Parse\_Response}(o)$\tcp*[l]{\small Extract Label and Confidence}
                Append $[(f.\text{id}, f.\text{timestamp}, c, p, \mathscr{m}, \text{label}, \zeta $),
                $ (\texttt{vram}, \texttt{latency}, \texttt{power})]$ to $\mathcal{L}$\;
            }
        }
    }
}

\tcp{\small Phase 3 - Threat Evaluation.}
\ForEach{entry $(f.\text{id}, f.\text{timestamp}, c, p, \mathscr{m}, \text{label}, \zeta, \dots) \in \mathcal{L}$}{
    \If{$\zeta \geq \tau$ \textbf{and} label = \texttt{Yes}}{
        {Save frame $f$ as an alert for human review}\;
    }
}
Export $\mathcal{L}$ as \ac{CSV}\;

\end{algorithm}

\item The model returns a textual response to the prompt for each frame. These responses are parsed to extract a binary label (\texttt{Yes}/\texttt{No}) and a confidence score $\zeta$, which are then stored in the log $\mathcal{L}$ alongside system-level statistics.

\end{enumerate}

\subsection{Phase 3 - Threat Evaluation}
{
To assess label-consistent retrieval, each affirmative response is associated with the video-level category used to formulate the query.
We employ the model's internal confidence $\zeta$ to filter and rank affirmative responses for alert review.
Only frames with a positive response and $\zeta \geq \tau$ are stored. In our experiments, $\tau$ is set separately for each model to the tenth-highest confidence among its positive responses.

The forensic specialization of \sol\ derives from this investigative workflow. Queries encode an operator-defined investigative need; each shortlisted output retains the flagged frame and its link to the source video; and the structured log records the frame identifier, source timestamp, query, normalized response, model, confidence, and computational measurements. A shortlisted output therefore becomes a traceable alert that an officer can review in the context of the original footage. The system supports rapid evidence triage and leaves the interpretation of the flagged material and any investigative decision to the human operator.
}

To better display our surveillance pipeline, we implemented a modular per-frame evaluation routine detailed in Algorithm~\ref{alg:vlm_pipeline}.
{Together, category-specific prompting, uniform sampling, confidence-based filtering, and structured logging constitute the deployment strategy examined in \textbf{RQ3}.}

\section{Experimental Protocol}
\label{sec:experiments}

To verify the effectiveness of \ac{VLM} in retrieving information from $f$, we defined an experimental protocol and quality metrics to parametrize the difficulty of the task and the processing performance of each models.

Each frame, along with the corresponding prompt, is fed into the selected \ac{VLM} in inference mode (no additional gradient computation nor fine tuning).
The models' language response is post-processed into one of three normalized categories: \textit{Yes}, \textit{No}, or \textit{Unclear}.
{The complete normalized response stream is used to calculate the reported metrics: \textit{Yes} is a positive response, while \textit{No} and \textit{Unclear} are non-alert responses. For operational use, a positive response is included in the alert shortlist only when $\zeta \geq \tau$. The score therefore acts as a within-model salience filter that prioritizes frames for human review; it is not interpreted as the probability that a crime occurred.}
Per-frame metadata is logged to structured \ac{CSV} and plain-text logs.

\subsection{Confidence-based Filtering}
{
Confidence is computed as the average of the maximum \texttt{softmax} probabilities over all generated tokens~\cite{petryk_2024_wacv}.
Specifically, let $\mathbf{s}_i$ be the raw logits produced at generation step $i$, and $\sigma(\mathbf{s}_i)$ the corresponding \texttt{softmax} distribution.
The confidence $\zeta$ for a response is defined as

\begin{equation}
\label{eq:confidence}
\zeta = \frac{1}{\mathcal{T}} \sum_{i=1}^{\mathcal{T}} \max_j \sigma(\mathbf{s}_i)_j,
\end{equation}

where $\mathcal{T}$ is the number of tokens generated (typically $\leq 5$).
This value summarizes the concentration of the model's token distributions while generating its response.
It is used to filter and rank flagged frames and, together with alert frequency and the validation results, to inspect the model's over-activation profile.
}

We release the code of \sol\ as open-source at~\cite{vlm_surv_repo} to promote the reproducibility of our results and encourage the readers to extend its deployment.

\section{Composite Model Utility Score (\ac{CMUS})}
\label{sec:cmus}

To assess the performance of \acp{VLM} in joint terms of detection robustness, efficiency, and semantic reliability, we introduce the \textbf{\acl{CMUS}}.
This metric integrates detection coverage, time efficiency, detection rate, and a refined hallucination index into a single, interpretable score, facilitating comparative assessment across models. The objective of the metric is to display the highest score possible for a fitting model.

\subsection{Scoring Terms}
\ac{CMUS} considers the following quantities for each model $\mathscr{m} \in \mathcal{M}$, where $\mathcal{M}$ is the set of the 4 models:
\begin{itemize}
    \item {$C_{\mathscr{m}}$: \textbf{Coverage} -- proportion of videos for which the model produces at least one positive response, averaged across the evaluated categories and prompt levels. Coverage measures retrieval breadth: a high value means that fewer videos are entirely missed during triage, but it does not imply that every positive response is correct.}
    \item $T_{\mathscr{m}}$: \textbf{Time Saved} -- relative reduction in processing time compared to the original duration of the video, averaged across the evaluated categories ($T_s$ values computed for each model $\mathscr{m}$).
    High values are preferred.
    \item $D_{\mathscr{m}}$: \textbf{Detection Rate} -- average proportion of frames receiving a positive response. Excessively high values can indicate hallucination.
    \item $H_{\mathscr{m}}$: \textbf{Refined Hallucination Index} -- a weighted sum over five detection buckets indicating the severity of the hallucination.
\end{itemize}

\subsubsection{Detection Coverage}
{
For each model, coverage is computed as the proportion of videos containing at least one positive response to the category-specific query.
This video-level quantity estimates whether the pipeline retrieves at least one potentially useful frame from each recording, which is the primary requirement of a triage system.
Coverage therefore complements the frame-level detection rate: the former measures how broadly the dataset is covered, whereas the latter measures how frequently a model produces positive responses within the analyzed videos.
}

\subsubsection{Temporal Efficiency Analysis}
To evaluate computational efficiency, we compared the average inference duration $T_{in}$ per model against the total actual playback time $T_{or}$ of the videos, as the latter offers a realistic baseline term for human or real-time monitoring systems.
We then computed the average inference time for each model by averaging over the three prompt complexity levels (easy, medium, and hard).

To quantify the proportional gain, we define the Time Saving as $T_s$ for model $\mathcal{M}$ as reported in Eq. \ref{eq:time_saved}:

\begin{equation}
T_s (\%) = \left( \frac{T_{or} - T_{in}}{T_{or}} \right) \times 100.
\label{eq:time_saved}
\end{equation}


\subsubsection{Detection Rate and Refined Hallucination Index}
{
Some general-purpose techniques, such as the method proposed by \textit{Li et al.}~\cite{li_2024_emnlp}, rely on uncertainty estimates to detect hallucinations in \acp{VLM} without requiring ground-truth annotations. Although efficient, these methods are not tailored to the operational constraints and forensic needs of surveillance scenarios. In such settings, false positives can significantly undermine analyst trust and delay critical decision-making, especially when reviewing large volumes of footage.

For each \texttt{Model}$\times$\texttt{Video}$\times$\texttt{Prompt} triplet, the detection rate is the proportion of analyzed frames for which the model returns \texttt{Yes}.
Because exhaustive frame-level ground truth is unavailable, we treat persistent affirmative activation as an operational proxy for hallucination risk. This proxy does not constitute direct proof of an incorrect response.
We group detection rates into five intervals and assign progressively stronger penalties:

\begin{enumerate}
    \item $<10\%$: selective activation and the lowest penalty;
    \item $10-40\%$: moderate activation with no penalty;
    \item $40-70\%$: increased risk of over-activation;
    \item $70-95\%$: high risk of persistent over-activation;
    \item $\geq95\%$: near-constant activation and the highest penalty.
\end{enumerate}

These intervals form the refined hallucination index and make its operational assumption explicit.

We model the resulting bucket-weighted score as in Eq.~\ref{eq:hallucination}:

\begin{equation}
\label{eq:hallucination}
H_{\mathscr{m}} = \frac{1}{N} \sum_{k=1}^{5} w_k \cdot B_k.
\end{equation}
where $B_k$ is the number of model-video-prompt triplets in bucket $k$, $N$ is the total number of triplets for that model, and the weights $w_k$ are set to $\{-0.5,\ 0.0,\ 0.5,\ 1.0,\ 1.5\}$ according to the intervals above.

In the crime-video evaluation, \(H_m\) is computed over the \(2{,}850\) video–prompt configurations. For the normal-video control, it is independently recomputed over the \(1{,}950\) video–category–prompt configurations using the same detection-rate intervals and bucket weights.

We assess this operational proxy through two complementary checks. A manually inspected human-reference subset evaluates whether model responses align with visually recognizable instances of the video-level target activity. A normal-video control group evaluates affirmative activation when none of the 13 target crime categories is expected.
}

\subsection{Final Score Definition}
The \ac{CMUS} in Eq. \ref{eq:cmus} integrates all four components using a weighted linear combination:
\begin{equation}
\label{eq:cmus}
\text{CMUS}_{\mathscr{m}} = \alpha \cdot C_{\mathscr{m}} + \beta \cdot T_{\mathscr{m}} + \gamma \cdot D_{\mathscr{m}} + \delta \cdot H_{\mathscr{m}}.
\end{equation}

\noindent For this experiment, we set:
\begin{itemize}
    \item $\alpha = 1.0$: Detection coverage is directly rewarded as it reflects the ability of the model to identify relevant activity across various scenarios.
    \item $\beta = 1.0$: Time savings are equally important, especially for real-time or resource-constrained systems, where faster inference enables scalability.
    \item $\gamma = - 0.5$: The detection rate is penalized moderately, as excessive detections may signal overgeneralization or hallucinations. However, occasional high activity may still be legitimate.
    \item $\delta = - 1.0$: The refined hallucination index receives full penalty weight, as hallucinated outputs pose a serious risk in forensic or security-sensitive applications.
\end{itemize}

This weighting scheme prioritizes models that are broadly effective and efficient, while strongly discouraging false or misleading detections.

\section{Results and Discussion}
\label{sec:results}
Unless otherwise stated, all models are evaluated under the same experimental conditions.

\begin{figure*}[t]
    \centering
    \subfloat[\scriptsize Robbery--LLaVA, $0.79$ conf. (medium)]{
        \includegraphics[width=0.30\textwidth]{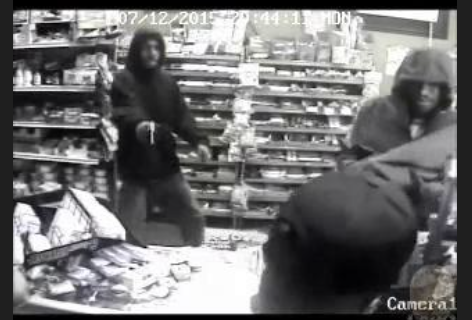}
        \label{fig:robbery_example}
    }\hfill
    \subfloat[\scriptsize Fighting--Qwen2.5-VL, $0.93$ conf. (easy)]{
        \includegraphics[width=0.30\textwidth]{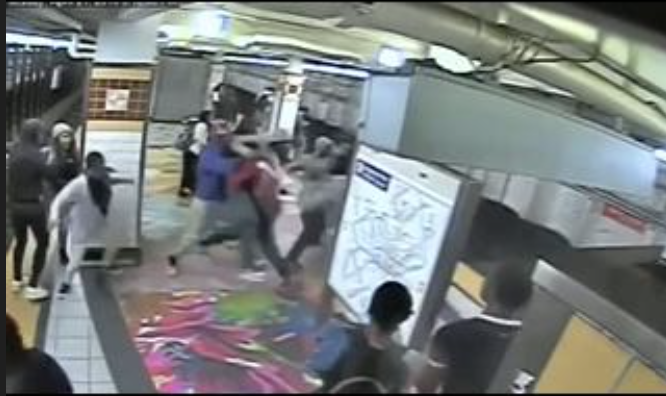}
        \label{fig:fighting_example}
    }\hfill
    \subfloat[\scriptsize Arson--InstructBLIP, $0.85$ conf. (hard)]{
        \includegraphics[width=0.30\textwidth]{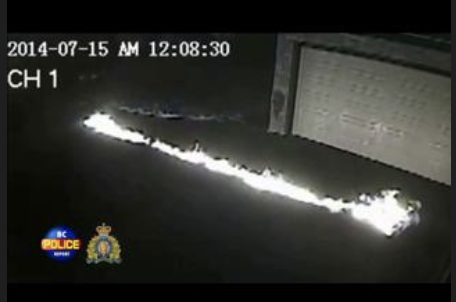}
        \label{fig:arson_example}
    }
    \caption{{Examples of successful detections from three models across different actions and prompt difficulties. Confidence values are internal generation scores used to filter and rank detections; they are not calibrated probabilities of correctness.}}
    \label{fig:qual_examples}
\end{figure*}

Fig.~\ref{fig:qual_examples} shows qualitative examples of successful frame-level detections.
Fig.~\ref{fig:robbery_example} shows a man holding a cashier at gunpoint, Fig.~\ref{fig:fighting_example} displays a brawl in a train/metro station, and Fig.~\ref{fig:arson_example} depicts a line of fire.
Each example is extracted from a successfully identified surveillance scenario, captured with different prompt difficulties, and processed by distinct models.

\subsection{Model-level Detection Performance}
{
To summarize the detection behavior of each \ac{VLM}, we computed the average coverage and per-frame detection rate across all categories and prompt difficulties.
Table~\ref{tab:model_category_performance} reports the category-wise averages for each model and the global performance summary.

Coverage represents the percentage of videos in a category for which at least one positive response (\texttt{Yes}) was produced.
The detection rate indicates the average ratio of frames marked as positive compared to the total number of frames analyzed.
Together, these metrics help distinguish between truly robust recognition and overactive or hallucinated detections.

\textbf{InstructBLIP} shows the highest overall rate, with a mean detection coverage of $99.08\%$ and a per-frame detection rate of $0.83$.
However, this high detection saturation likely stems from a tendency to over-predict positives, potentially inflating performance at the cost of hallucinations.
\textbf{DeepSeek-VL} ranks second, averaging $95.6\%$ coverage and a $0.62$ detection rate, suggesting frequent affirmative activation.

\textbf{LLaVA} and \textbf{Qwen2.5-VL}, in contrast, display lower detection rates ($0.27$ and $0.16$ respectively), yet this conservative behavior likely avoids unnecessary hallucinations and better reflects real-world constraints.
\textit{Qwen2.5-VL} in particular exhibits selective detection patterns while maintaining respectable detection coverage ($66.9\%$), suggesting a potential balance between efficiency and semantic caution.

It is important to note that these metrics should not be considered without accounting for hallucinations.
Models with high detection counts may also exhibit elevated false-positive rates.
These results address \textbf{RQ1} by showing the extent to which \acp{VLM} can retrieve surveillance evidence and inform \textbf{RQ2} by exposing the different operational trade-offs among the evaluated architectures.
}

\begin{table*}[!t]
\centering
\caption{{Average coverage (\%) and detection rate per model and category. Values are averaged across the three prompt difficulty levels (easy, medium, hard). Normal videos are excluded.}}
\label{tab:model_category_performance}
\begin{tabular}{l|cc|cc|cc|cc}
\toprule
& \multicolumn{2}{c|}{\textbf{DeepSeek-VL}} & \multicolumn{2}{c|}{\textbf{InstructBLIP}} & \multicolumn{2}{c|}{\textbf{LLaVA}} & \multicolumn{2}{c}{\textbf{Qwen2.5-VL}} \\
\textbf{Category} & Cov. (\%) & Det. Rate & Cov. (\%) & Det. Rate & Cov. (\%) & Det. Rate & Cov. (\%) & Det. Rate \\
\midrule
Abuse & $98.67$ & $0.8353$  & $98.00$ & $0.8582$ & $50.67$ & $0.1652$ & $32.67$ & $0.0292$ \\
Arrest & $100.00$ & $0.7011$  & $100.00$ & $0.8283$ & $86.00$ & $0.3258$ & $80.00$ & $0.2454$ \\
Arson & $96.67$ & $0.4096$  & $100.00$ & $0.6046$ & $62.00$ & $0.0969$ & $71.33$ & $0.0800$ \\
Assault & $96.67$ & $0.6202$  & $98.67$ & $0.8525$ & $70.67$ & $0.2162$ & $59.33$ & $0.1327$ \\
Burglary & $95.33$ & $0.6059$  & $100.00$ & $0.9717$ & $89.00$ & $0.4351$ & $79.00$ & $0.2271$ \\
Explosion & $96.00$ & $0.4323$  & $97.33$ & $0.5994$ & $68.00$ & $0.2079$ & $86.67$ & $0.3053$ \\
Fighting & $100.00$ & $0.6081$  & $100.00$ & $0.8681$ & $92.00$ & $0.2198$ & $87.33$ & $0.2106$ \\
RoadAccidents & $93.33$ & $0.6439$  & $98.67$ & $0.9226$ & $84.89$ & $0.4200$ & $63.34$ & $0.2160$ \\
Robbery & $99.78$ & $0.7359$  & $100.00$ & $0.9695$ & $88.89$ & $0.3778$ & $87.33$ & $0.2710$ \\
Shooting & $71.33$ & $0.2116$  & $96.67$ & $0.6216$ & $35.33$ & $0.0133$ & $23.33$ & $0.0143$ \\
Shoplifting & $98.00$ & $0.8134$  & $100.00$ & $0.9663$ & $90.67$ & $0.3923$ & $71.33$ & $0.1002$ \\
Stealing & $97.33$ & $0.6363$  & $99.33$ & $0.8988$ & $62.67$ & $0.1933$ & $75.33$ & $0.1435$ \\
Vandalism & $100.00$ & $0.7703$  & $99.33$ & $0.8335$ & $82.00$ & $0.4395$ & $52.67$ & $0.0769$ \\
\midrule
\textbf{Average} & $95.62$ & $0.6172$ & \textbf{$99.08$} & \textbf{$0.8304$} & $74.06$ & $0.2695$ & $66.90$ & $0.1579$ \\
\bottomrule
\end{tabular}
\end{table*}

\subsection{Hallucination occurrence}

{
Table~\ref{tab:hallucination_buckets} summarizes the $2{,}850$ video--prompt pairs assigned to each detection-rate band for every model. \textbf{Qwen2.5-VL} exhibits the lowest hallucination footprint: $1{,}591$ pairs fall below the $10\%$ threshold and only $12$ reach the $\geq95\%$ band, resulting in $H_m=-0.1598$.
\textbf{InstructBLIP} shows the strongest persistent affirmative activation, with $2{,}032$ pairs in the $\geq95\%$ band and only $106$ below $10\%$. It consequently obtains the highest hallucination index, $H_m=1.2030$.

\textbf{DeepSeek-VL} also presents substantial over-activation: $650$ pairs fall in the $70$--$95\%$ band and $880$ in the $\geq95\%$ band, yielding $H_m=0.7063$.
\textbf{LLaVA} displays a more selective profile, with $1{,}223$ pairs below $10\%$ and $123$ in the highest band, resulting in $H_m=0.0609$.

These trends emphasize the necessity of interpreting isolated ``Yes'' detections within the broader activation pattern of each model.
Persistent positive responses across low-activity, occluded, or otherwise ambiguous frames may indicate over-activation and limited selectivity.
}

\begin{table}[!t]
\centering
\caption{Number of video--prompt pairs falling into detection-rate bands.}

\label{tab:hallucination_buckets}
\renewcommand{\arraystretch}{1.1}
\resizebox{\columnwidth}{!}{%
\begin{tabular}{l|ccccc|c}
\hline
\textbf{Model} & \textbf{$<10\%$} & \textbf{$10-40\%$} & \textbf{$40-70\%$} & \textbf{$70-95\%$} & \textbf{$\geq95 \%$} & $H_m$\\
\hline
Qwen2.5-VL     & $1591$ & $753$ & $344$ & $150$ & $12$  & $\bf{-0.1598}$ \\
InstructBLIP   & $106$ & $182$  & $193$  & $337$  & $2032$ & $1.2030$ \\
DeepSeek-VL    & $390$  & $454$  & $476$  & $650$ & $880$ & $0.7063$ \\
LLaVA          & $1223$ & $658$ & $491$ & $355$ & $123$ & $0.0609$  \\
\hline
\end{tabular}
}
\end{table}

{This analysis supports a nuanced interpretation of affirmative outputs in \ac{VLM}-based surveillance: robust detectors balance detection breadth and semantic control.
While \textit{DeepSeek-VL} ensures coverage, \textit{Qwen2.5-VL} remains the most judicious in limiting persistent over-activation, making it preferable for high-stakes forensic applications where hallucination undermines credibility.}

\subsection{Inspection time speed-up}
Table~\ref{tab:time_savings_summary} reports the mean processing time across the 13 categories, whose videos have an average duration of $155.77$ minutes. All models complete inference faster than raw playback. InstructBLIP and Qwen2.5-VL save $90.96\%$ and $85.32\%$ of the inspection time, while DeepSeek-VL and LLaVA save $74.13\%$ and $64.59\%$. {Category-level values are provided in the supplementary material.} 

\begin{table}[!t]
\centering
\scriptsize

\caption{Mean processing time and time saved across categories.}
\label{tab:time_savings_summary}
\begin{tabular}{lcc}
\toprule
\textbf{Model} & \textbf{Processing time (min)} & \textbf{Time saved (\%)} \\
\midrule
DeepSeek-VL & $40.29$ & $74.13$ \\
InstructBLIP & $14.09$ & $90.96$ \\
LLaVA & $55.17$ & $64.59$ \\
Qwen2.5-VL & $22.82$ & $85.32$ \\
\bottomrule
\end{tabular}
\end{table}

\subsection{\ac{CMUS} Scores}
Table~\ref{tab:cmus_summary} presents both the normalized inputs and the resulting \ac{CMUS} values for each model after applying Eq.~\ref{eq:cmus}.

\begin{table}[!t]
\centering
\caption{Normalized \ac{CMUS} inputs and resulting scores.}
\label{tab:cmus_summary}
\renewcommand{\arraystretch}{1.2}
\resizebox{\columnwidth}{!}{%
\begin{tabular}{l|cccc|c}
\hline
\textbf{Model} & $C_m$ & $T_m$ & $D_m$ & $H_m$ & \textbf{CMUS score} \\
\hline
Qwen2.5-VL & $0.6690$ & $0.8532$ & $0.1579$ & $-0.1598$ & $\mathbf{1.6031}$ \\
InstructBLIP & $0.9908$ & $0.9096$ & $0.8304$ & $1.2030$ & $0.2822$ \\
DeepSeek-VL & $0.9562$ & $0.7413$ & $0.6172$ & $0.7063$ & $0.6826$ \\
LLaVA & $0.7406$ & $0.6459$ & $0.2695$ & $0.0609$ & $1.1909$ \\
\hline
\end{tabular}
}
\end{table}

These results illustrate the trade-offs between detection aggressiveness, semantic control, and computational efficiency:

\begin{itemize}
    \item \textbf{Qwen2.5-VL} obtains the highest \ac{CMUS} score ($1.6031$). Its low detection rate ($0.1579$) and negative hallucination index ($H_{\mathscr{m}}=-0.1598$) offset its lower coverage.
    \item \textbf{LLaVA} ranks second with a \ac{CMUS} score of $1.1909$, supported by a low hallucination index ($H_{\mathscr{m}}=0.0609$) and a moderate detection rate.
    \item \textbf{DeepSeek-VL} and \textbf{InstructBLIP} obtain scores of $0.6826$ and $0.2822$, respectively. Their high coverage is offset by frequent affirmative outputs and higher hallucination indices.
\end{itemize}

The \ac{CMUS} thus provides an operational aggregate for comparative model selection in security, surveillance, and forensic contexts. Its ranking is consistent with the manually referenced evaluation and the normal-video control.

\subsection{Human-Reference Validation}
\label{sec:human_reference_validation}

To validate the reference-free evaluation, we manually inspected five videos from each of the 13 crime categories, for a total of 65 videos and $5{,}909$ analyzed frames. Each frame was assigned a binary human reference indicating whether activity related to the corresponding video-level category was visually recognizable. The resulting reference contains $2{,}342$ positive frames and $3{,}567$ negative frames.

Model responses were compared with this reference using the same output-normalization rule adopted by \sol. A \textit{Yes} response is a positive prediction; \textit{No} and \textit{Unclear} responses are non-alert predictions. This rule also covers non-binary variants such as ``uncertain,'' which are normalized as \textit{Unclear}. The human reference is repeated across the three prompt formulations, producing $17{,}727$ frame--prompt decisions per model. We report the true-positive rate $\mathrm{TPR}=\mathrm{TP}/(\mathrm{TP}+\mathrm{FN})$ and false-positive rate $\mathrm{FPR}=\mathrm{FP}/(\mathrm{FP}+\mathrm{TN})$, with $\mathrm{TP}$ and $\mathrm{TN}$ the number of true positives and negatives, $\mathrm{FP}$ and $\mathrm{FN}$ false positives and negatives, respectively.

Table~\ref{tab:human_reference_validation} reports the aggregate false-positive and true-positive rates across the three prompt formulations.

\begin{table}[!t]

\centering
\caption{Human-reference false-positive and true-positive rates aggregated across the three prompt formulations.}
\label{tab:human_reference_validation}
\begin{tabular}{lcc}
\toprule
\textbf{Model} & $\mathrm{FPR}_{\mathrm{human}}$ & $\mathrm{TPR}_{\mathrm{human}}$ \\
\midrule
Qwen2.5-VL   & $0.0644$ & $0.2535$ \\
LLaVA        & $0.1781$ & $0.4073$ \\
DeepSeek-VL  & $0.5497$ & $0.7832$ \\
InstructBLIP & $0.8195$ & $0.9471$ \\
\bottomrule
\end{tabular}
\end{table}

The crime-video \(H_m\), the independently computed normal-video \(H_m\), and \(\mathrm{FPR}_{\mathrm{human}}\) yield the same model ordering: Qwen2.5-VL, LLaVA, DeepSeek-VL, and InstructBLIP. This is also the ordering produced by \ac{CMUS} in both the primary crime-video evaluation and the normal-video control. The supervised human-reference results and the normal-video experiment are therefore consistent with the reference-free ranking. Models with a higher $H_m$ also produce more affirmative responses on normal footage and a higher false-positive rate against the human reference. This agreement supports the use of persistent affirmative activation as an operational proxy for hallucination risk and supports the comparative \ac{CMUS} result within the evaluated setting.

The human-reference results also expose the associated sensitivity trade-off. Qwen2.5-VL is the most selective model, with an FPR of $0.0644$ and a TPR of $0.2535$. InstructBLIP reaches a TPR of $0.9471$, but its FPR of $0.8195$ indicates pervasive affirmative activation. DeepSeek-VL occupies an intermediate high-activation profile, while LLaVA retains lower false-alert activation with moderate sensitivity.

LLaVA produced ``uncertain'' responses for $4{,}720$ of the $5{,}909$ frames ($79.88\%$) under the medium prompt. These outputs were normalized as \textit{Unclear} and did not trigger alerts. For this prompt, LLaVA consequently obtained an FPR of $0.1365$ and a TPR of $0.1699$. This behavior indicates sensitivity to prompt formulation: abstaining limits false alerts but also reduces the number of recognizable target frames retrieved.

The successful cases in Fig.~\ref{fig:qual_examples} illustrate frames in which the target activity is visually recognizable. The manual inspection provides qualitative context for the remaining disagreements. They commonly involve ambiguous or overlapping activities, partial occlusion, events occurring outside the camera view, and acquisition artifacts such as frozen frames. Discontinuous positive responses across adjacent frames are also compatible with the independent frame-level design. These cases limit exact temporal localization but do not prevent the system from retrieving a useful approximation of the relevant interval for analyst review.

\subsection{Control Group: Normal Videos}
\label{sec:normal_videos}
The control group contains 50 normal videos and follows the same sampling and inference protocol used for the crime videos. Each video is queried separately for each of the 13 crime categories using the easy, medium, and hard prompt formulations. This produces 1,950 video--category--prompt configurations per model. Since these videos contain none of the target crime categories, affirmative responses are treated as false alerts in this control analysis. Table~\ref{tab:normal_video_control} reports the resulting metrics.

\begin{table}[!t]
\centering
\caption{Control-group evaluation on 50 normal videos queried for all 13 crime categories at three prompt levels. All reported quantities are computed specifically on the control group.}
\label{tab:normal_video_control}
\resizebox{\columnwidth}{!}{%
\begin{tabular}{l|cccc|c}
\hline
\textbf{Model} & $C_m$ & $T_m$ & $D_m$ & $H_m$ & \textbf{CMUS score} \\
\hline
Qwen2.5-VL & $0.0154$ & $0.8421$ & $0.0007$ & $-0.5000$ & $\bf{1.3571}$ \\
InstructBLIP & $0.6205$ & $0.9069$ & $0.4191$ & $0.29$ & $1.0279$ \\
DeepSeek-VL & $0.3877$ & $0.7266$ & $0.1711$ & $-0.1767$ & $1.2054$ \\
LLaVA & $0.0959$ & $0.6362$ & $0.0143$ & $-0.4867$ & $1.2116$ \\
\hline
\end{tabular}
}
\end{table}

{ The control results reproduce the ordering defined by $H_m$. Qwen2.5-VL records the lowest coverage and detection rate, followed by LLaVA, DeepSeek-VL, and InstructBLIP.} The \ac{CMUS} ranking obtained on the normal-video control also reproduces the primary crime-video ranking, and the same ordering is observed for the human-reference FPR in Table~\ref{tab:human_reference_validation}. These results provide a negative-control check of the refined hallucination index and an independent consistency check of the reference-free \ac{CMUS} comparison.

For the strict category-specific evaluation, every affirmative response on these videos is counted as a false alert. Some flagged frames may still depict unusual behavior or contextual elements that merit operator attention, despite not matching the queried crime category. This operational plausibility is retained in the qualitative interpretation, while the quantitative control keeps the stricter false-alert assignment.

\subsection{Computational performance}
\begin{figure}[t]
    \centering
    \includegraphics[width=0.8\linewidth, trim={0 0 0 1cm},clip]{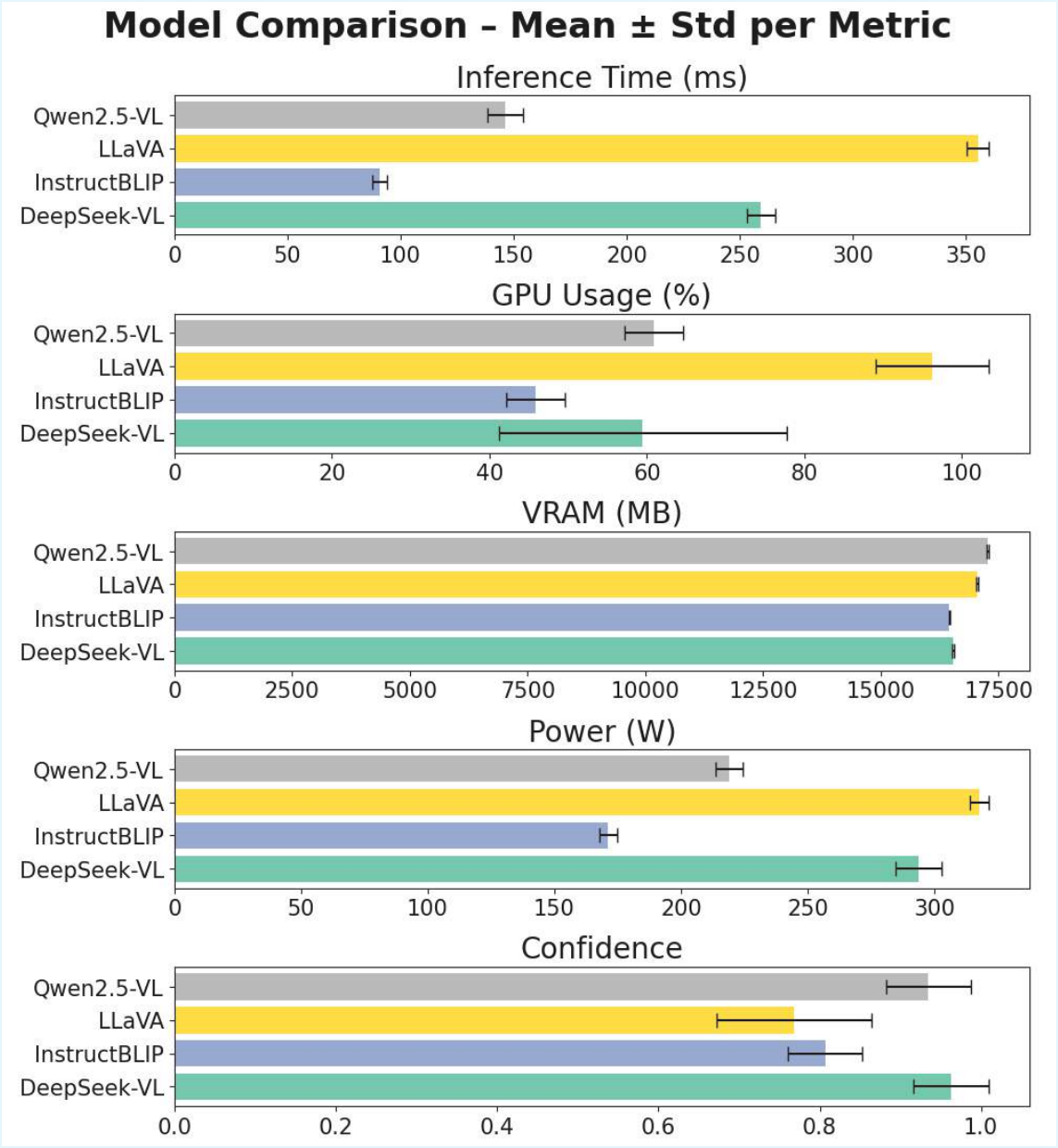}
    \caption{Mean and standard deviation of each metric across all models, prompts, and categories. Each bar aggregates over $150{,}000$ frames.}
    \label{fig:metric_summary}
\end{figure}

{Figure~\ref{fig:metric_summary} summarizes the computational profile obtained from more than $150{,}000$ frame-level inference records per model. InstructBLIP is the fastest model ($\sim90$ ms/frame) and has the lowest sampled power draw ($\sim170$ W); Qwen2.5-VL follows at $\sim145$ ms/frame and $\sim215$ W. DeepSeek-VL and LLaVA require approximately $260$ and $350$ ms/frame and draw about $295$ and $310$ W, respectively. All four models use approximately $16.5$--$17.5$ GB of \ac{VRAM}. Confidence values are used only to filter and rank affirmative responses and are not calibrated correctness probabilities. Detailed frame-level profiles and normalized comparisons are reported in the supplementary material.}

\subsection{Answers to the Research Questions}
\label{sec:rq_answers}

The experimental evidence allows us to answer the research questions introduced in Section~\ref{sec:intro} within the scope and assumptions stated in Section~\ref{sec:scope_assumptions}.

\begin{itemize}
    \item[\textbf{RQ1.}] \acp{VLM} can support surveillance analysis by retrieving frames related to an operator-defined query and substantially reducing the footage that requires direct inspection. Their usefulness nevertheless depends on model selectivity and the ambiguity of the source video.
    \item[\textbf{RQ2.}] No architecture dominates every operational dimension. The comparison instead identifies different trade-offs among retrieval breadth, output selectivity, inference time, and hardware demand, allowing the model to be selected according to the deployment priority.
    \item[\textbf{RQ3.}] Consistent integration requires category-specific prompts, controlled frame sampling, normalized binary outputs, confidence-based filtering, and structured logging of both detections and computational measurements.
    \item[\textbf{RQ4.}] Reference-free comparison is feasible through the joint interpretation of coverage, detection rate, hallucination risk, and temporal efficiency. Across $5{,}909$ manually referenced frames and the 50-video normal control, $H_m$, the normal-video detection rate, and the human-reference FPR produce the same model ordering as \ac{CMUS}. The reference-free, supervised, and control experiments therefore provide consistent comparative results within the evaluated benchmark.
\end{itemize}

\section{Limitations and Future Work}
\label{sec:future}

{
The evaluation inherits the intrinsic limitations of the \ac{UCF}-Crime dataset.
Its labels are assigned at video level and do not describe individual frames. The labeled activity may be ambiguous, overlap with other categories, occur outside the camera view, or be obscured by poor visibility, motion blur, occlusion, or frozen frames.
The manual human-reference subset covers 65 videos and $5{,}909$ frames, while the normal-video control covers 50 additional videos. These experiments provide complementary validation, but they do not turn the material into an exhaustive objective frame-level ground truth. Responses such as ``uncertain'' are treated as non-alert outcomes; their frequency, particularly for LLaVA under the medium prompt, also shows that the measured performance depends on prompt formulation and response normalization.

The current pipeline analyzes frames independently.
This design supports modular and efficient triage, but it does not model temporal continuity or guarantee continuous localization of an event.
The experiments also use a single workstation and off-the-shelf models without domain-specific fine-tuning; consequently, the reported hardware profile is specific to the evaluated configuration.
In particular, the \ac{NVML} measurements describe sampled power draw in W. They do not represent integrated energy consumption in Wh.

The same pipeline can support three operational settings. In post-incident triage, body-cam~\cite{chen_2019_cvpr} or fixed-camera footage is processed offline to retrieve frames related to an investigative query. Central evidence repositories can use higher-capacity models and return stored frames, timestamps, and response logs for analyst review. Compact models can also support edge or \ac{UAV} deployments, where lightweight vision systems are already practical~\cite{mehta_2022_iclr}. Prompt selection, model choice, and alert filtering can be configured for each hardware and investigative context; the flagged material remains subject to human assessment.

Future work may investigate lightweight temporal fusion, adaptive frame sampling, domain-specific adaptation, model compression, and controlled trials in the deployment settings described above while preserving the reference-free and operationally simple nature of the framework.
}

\section{Conclusion}
\label{sec:conclusion}
{
This work introduced \sol, a modular pipeline for prompt-based, per-frame surveillance analysis with 7B-parameter \acp{VLM}. Four models were evaluated across 13 crime categories and three prompt complexities using detection coverage, time saved, detection rate, and the hallucination index, combined through \ac{CMUS}. The framework retains the frames, timestamps, normalized responses, and computational logs needed for human inspection.

Qwen2.5-VL achieved the highest \ac{CMUS} ($1.6031$), followed by LLaVA ($1.1909$), reflecting their more selective activation profiles. DeepSeek-VL and InstructBLIP reached broader coverage but produced substantially more affirmative responses, leading to lower scores. All models processed frames faster than the configured $1$ Hz sampling interval, with average time savings from $64.59\%$ to $90.96\%$.

The same model ordering was obtained from $H_m$, the normal-video control, human-reference FPR, and both crime- and normal-video \ac{CMUS} evaluations. This agreement supports the comparative use of the reference-free score within the evaluated setting. The results position \sol\ as a practical triage and benchmarking framework whose alerts guide subsequent human assessment; its current limits remain the absence of exhaustive frame-level labels, independent-frame processing, and hardware-specific profiling.
}

\balance
\bibliographystyle{IEEEtran}
\bibliography{references}

\end{document}


\title{Supplementary Material for ``VESSI - VLM-Enhanced Support for Surveillance and Investigations''}
\author{Saverio Cavasin, Pietro Tedeschi, Mattia Tamiazzo, Alessandro Brighente, Simone Milani, and Mauro Conti}
\markboth{Supplementary Material for IEEE Transactions on Information Forensics and Security}{Cavasin \MakeLowercase{\textit{et al.}}: VESSI Supplementary Material}
\maketitle

\appendices

\renewcommand{\thetable}{S\arabic{table}}
\renewcommand{\thefigure}{S\arabic{figure}}
\renewcommand{\theequation}{S\arabic{equation}}

\section{Supplementary Technical and Reproducibility Details}
\label{app:technical_configuration}

\subsection{Hardware Platform}
We implemented \sol\ on a workstation equipped with a 24-core Intel Core i9-14900KF processor with up to $6.0$~GHz clock frequency, $62$~GB of \ac{DRAM}, and an NVIDIA GeForce RTX 4090 \ac{GPU} with $24$~GB of \ac{VRAM}.
The workstation runs Ubuntu 20.04 LTS with NVIDIA driver version 535.183.01 and \ac{CUDA} support.

\subsection{Software Environments}
The implementation uses two distinct \texttt{conda} environments because DeepSeek-VL requires a different library configuration from the other three models. This separation also prevents package conflicts:

\begin{itemize}
    \item \textbf{InstructBLIP, LLaVA, and Qwen2.5-VL Environment.} Configured with Python 3.10.13~\cite{python-3.10.13}, Transformers 4.49.0.dev0~\cite{transformers-4.49.0-dev0}, PyTorch 2.1.2~\cite{pytorch-2.1.2}, torchvision 0.16.2~\cite{torchvision-0.16.2}, and llava 1.2.2.post1~\cite{llava-1.2.2-post1}, which is specifically required for the LLaVA-1.5 implementation.
    \item \textbf{DeepSeek Environment.} Configured with Python 3.10.16 (same branch as 3.10.13), deepseek-vl 1.0.0~\cite{deepseek-vl-1.0.0}, Transformers 4.49.0~\cite{transformers-4.49.0}, PyTorch 2.6.0~\cite{pytorch-2.6.0}, and torchvision 0.21.0~\cite{torchvision-0.21.0}.
\end{itemize}

Both environments relied on PyTorch \ac{GPU} builds compiled for CUDA 12.1. In \textbf{Phase 2}, InstructBLIP, LLaVA, Qwen2.5-VL, and DeepSeek-VL were loaded through the Hugging Face \texttt{transformers} \ac{API}~\cite{wolf-etal-2020}, specifically using the \texttt{AutoProcessor.from\_pretrained()} and \texttt{AutoModelForCausalLM.from\_pretrained()} interfaces. The option \texttt{trust\_remote\_code=True} was enabled for Qwen2.5-VL and DeepSeek-VL.

\subsection{Custom \ac{NVML} Logging Module}
{We developed a custom logging module using \ac{NVML} via the \texttt{pynvml} Python bindings~\cite{pynvml-11.0.0}, which provides access to the \ac{NVML}, a C-based \ac{API} for monitoring \ac{GPU} state (e.g., utilization, memory, power, temperature)~\cite{nvml}. We used this library because it enables fine-grained querying of \ac{GPU} metrics that are not reliably available through high-level frameworks such as PyTorch. The \texttt{nvmlDeviceGetPowerUsage} call returns an instantaneous power sample at the time of measurement. Accordingly, the logged values characterize sampled power draw in W and must not be interpreted as energy consumption in Wh, which would require integration over a defined interval.}

Table~\ref{tab:log_fields} lists the metrics recorded for each inference step.
Frames associated with positive responses satisfying the confidence threshold are retained for subsequent inspection and ranked by confidence for qualitative evaluation.

\begin{table}[ht]
\caption{Logged metrics per frame.}

\label{tab:log_fields}
\centering
\renewcommand{\arraystretch}{1.2}
\begin{tabular}{|l|l|}
\hline
\textbf{Field} & \textbf{Description} \\
\hline
Frame ID            & Unique frame index \\
Timestamp (s)       & Relative time since start \\
Prompt              & Input natural language query \\
Model               & Model name (e.g., InstructBLIP) \\
Tokens              & Prompt token count \\
Readability         & \textit{Flesch-Kincaid} grade level \\
\ac{GPU} Usage (\%) & \ac{GPU} utilization \\
\ac{VRAM} (MB)      & Memory used on \ac{GPU} \\
{Power (W)} & {Instantaneous sampled power draw} \\
Inference Time (ms) & Time per inference \\
Confidence          & Model output confidence score \\
\hline
\end{tabular}
\end{table}

An example \ac{GPU} monitoring output during \textit{InstructBLIP} inference is shown in Table~\ref{tab:instructblip_arson_logging}. During these inference steps, the model process uses approximately $16.5$ GB of \ac{VRAM}. The associated prompt is: ``\textit{Could a human observer reasonably conclude that this frame depicts arson-like behavior or consequences? Answer Yes or No.}''

\begin{table}[!ht]
\scriptsize
\centering
\caption{Example \ac{GPU} monitoring entries for the \texttt{Arson} class using \textit{InstructBLIP} with a hard prompt.}
\label{tab:instructblip_arson_logging}
\setlength{\tabcolsep}{2.5pt}
\begin{tabular}{|c|c|c|c|c|c|}
\hline
\textbf{Frame} & \textbf{Time (s)} & \textbf{Model} & \textbf{\ac{VRAM} (MB)} & \textbf{Power (W)} & \textbf{Conf.} \\
\hline
$30$ & $1.00$ & InstructBLIP & $16,444.94$ & $174.87$ & $0.78$ \\
$60$ & $2.00$ & InstructBLIP & $16,450.94$ & $167.93$ & $0.78$ \\
$90$ & $3.00$ & InstructBLIP & $16,436.94$ & $167.32$ & $0.78$ \\
\hline
\end{tabular}
\vspace{2pt}

\parbox{0.97\columnwidth}{\scriptsize\textit{Prompt:} ``Could a human observer reasonably conclude that this frame depicts arson-like behavior or consequences? Answer Yes or No.''}
\end{table}

\subsection{Prompt Readability Assessment}
To quantitatively assess the semantic complexity of the prompts, we approximated the \textit{Flesch-Kincaid Grade Level} for each of the three template types (easy, medium, hard).
This metric estimates the U.S. school grade level required to understand a sentence, based on the ratio of words to sentences and the ratio of syllables to words.
Although standard readability libraries (e.g., \texttt{textstat}) are commonly used, we implemented a lightweight, dependency-free version based on the classic \textit{Flesch-Kincaid} equation~\cite{kincaid_1975}.
Let $\Omega$ be the number of words, $\Phi$ the number of sentences, $\Upsilon$ the number of syllables, $a$ and $b$ the weights influencing the ratios, and $c$ the bias term.
We denote the \textit{Flesch-Kincaid} grade as $FK$ in Eq.~\ref{fk_grade_general} as follows:

\begin{equation}
\label{fk_grade_general}
FK = a \cdot \frac{\Omega}{\Phi} + b \cdot \frac{\Upsilon}{\Omega} + c .
\end{equation}

The coefficients set as $a = 0.39$, $b = 11.8$, and $c = -15.59$ are obtained through linear regression on a dataset of U.S. Navy technical documents~\cite{kincaid_1975} and instructional materials, to align the output of the formula with human-assessed grade-level readability.
The first term $a$ models syntactic complexity via sentence length, the second term $b$ captures lexical difficulty through syllable density, and the bias term $c$ calibrates the scale to match the U.S. education system's grade levels.

The resulting readability scores for a generic category are:

\begin{itemize}
    \item \textbf{Easy prompt:} $FK = 1.47$---Simple and binary, designed for direct grounding of visual concepts.
    \item \textbf{Medium prompt:} $FK = 6.28$---Introduces interpretive phrasing, increasing semantic ambiguity.
    \item \textbf{Hard prompt:} $FK = 11.09$---Involves indirect reasoning and abstract language, challenging model inference.
\end{itemize}

\subsection{Detailed Computational Profiles}
\label{app:computational_profiles}
\begin{figure*}[t]
    \centering
    \subfloat[Inference Time]{\includegraphics[width=0.31\textwidth, trim={0 0 0 3cm},clip]{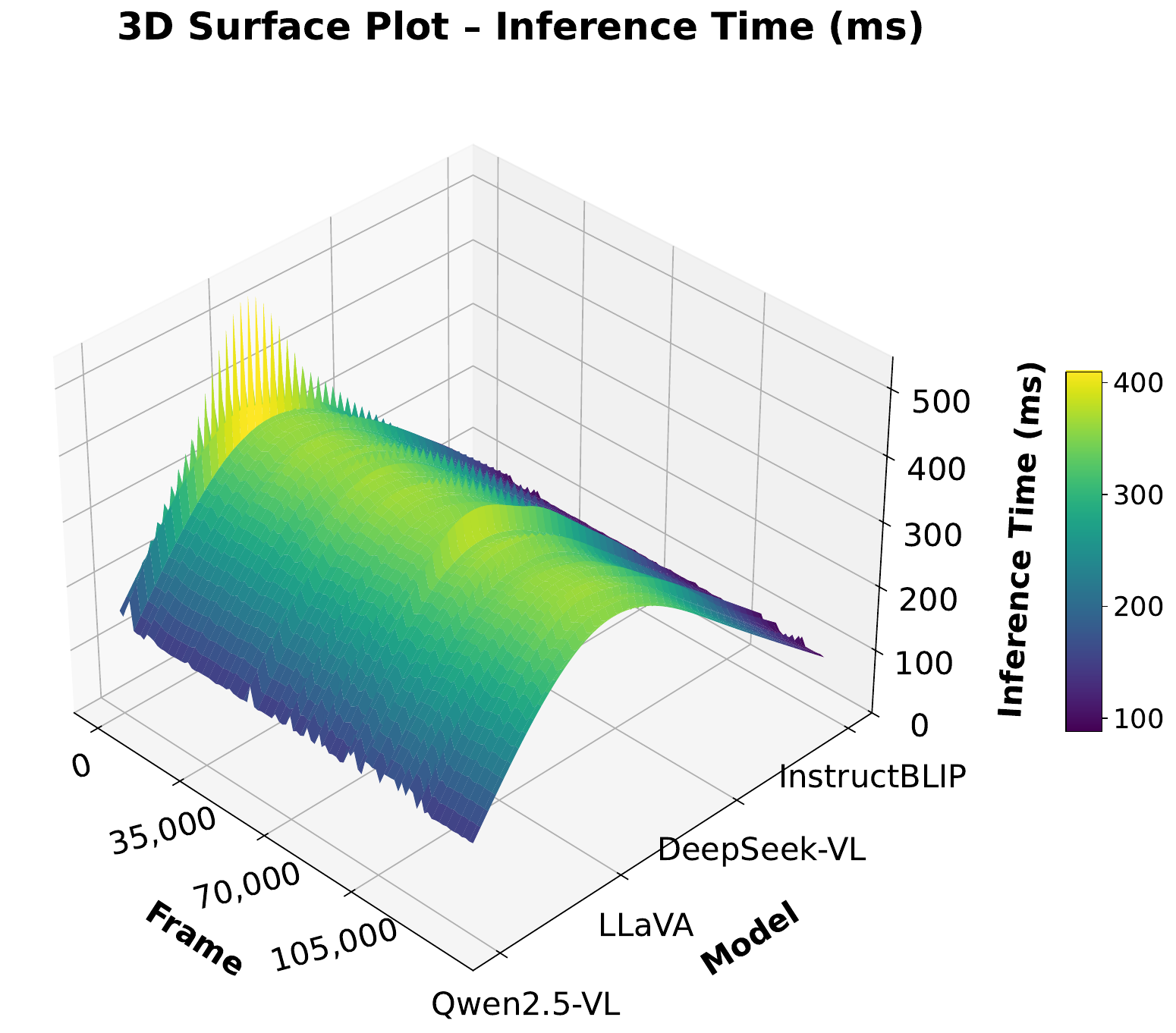}\label{fig:3d_time}}\hfill
    \subfloat[VRAM Consumption]{\includegraphics[width=0.31\textwidth, trim={0 0 0 3cm},clip]{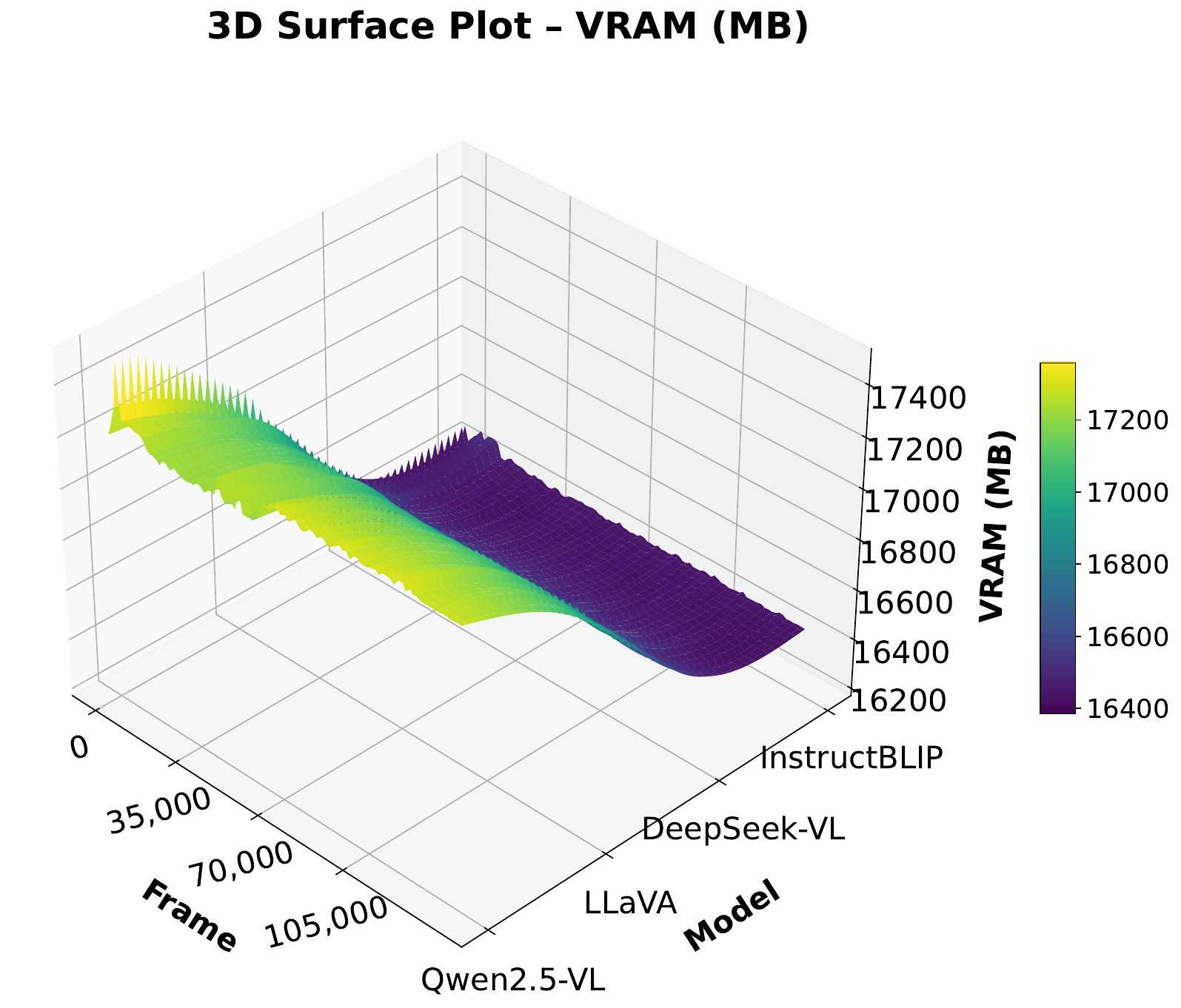}\label{fig:3d_vram}}\hfill
    \subfloat[GPU Utilization]{\includegraphics[width=0.31\textwidth, trim={0 0 0 3cm},clip]{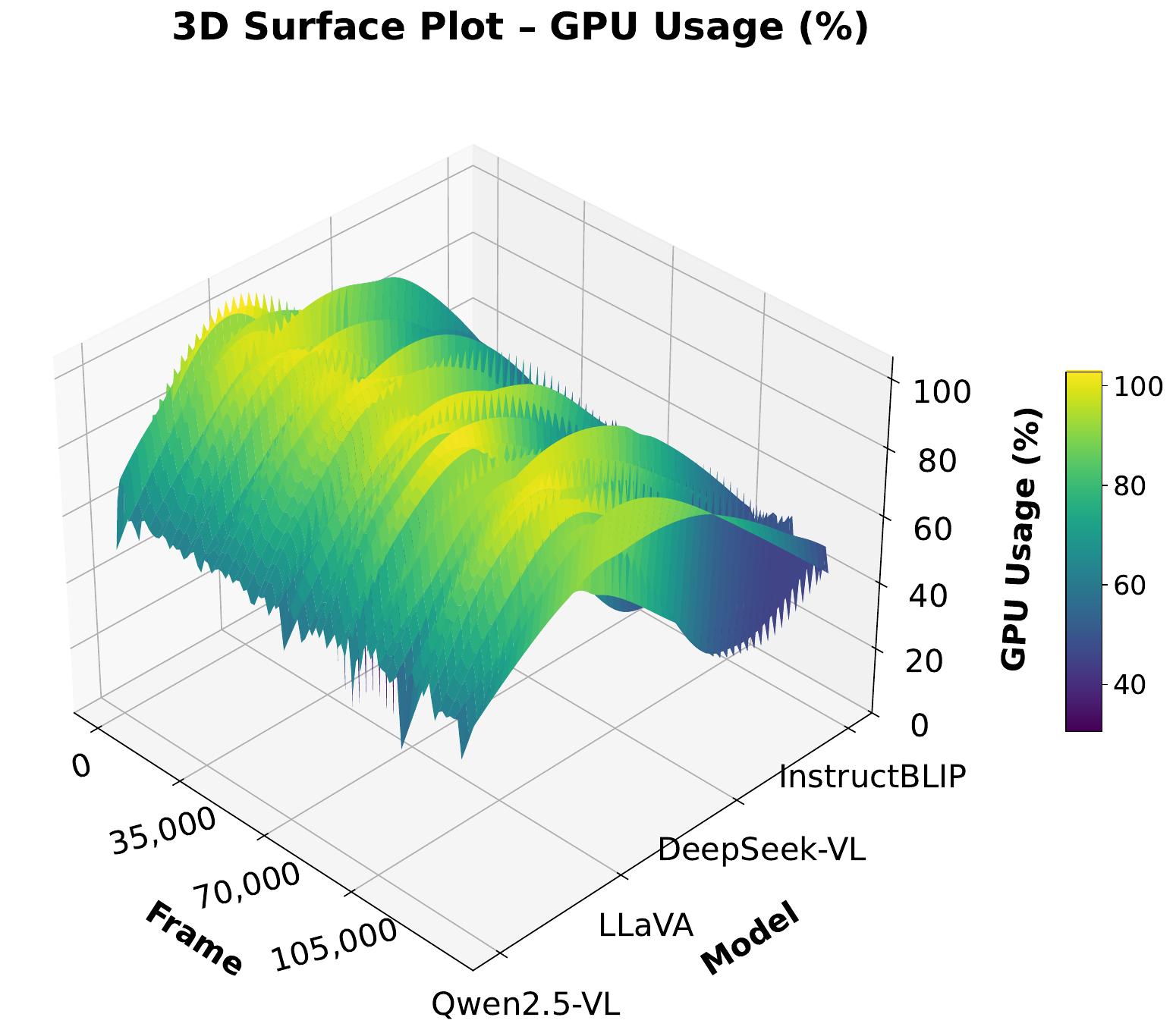}\label{fig:3d_gpu}}
    \par\medskip
    \subfloat[Power Draw]{\includegraphics[width=0.31\textwidth, trim={0 0 0 3cm},clip]{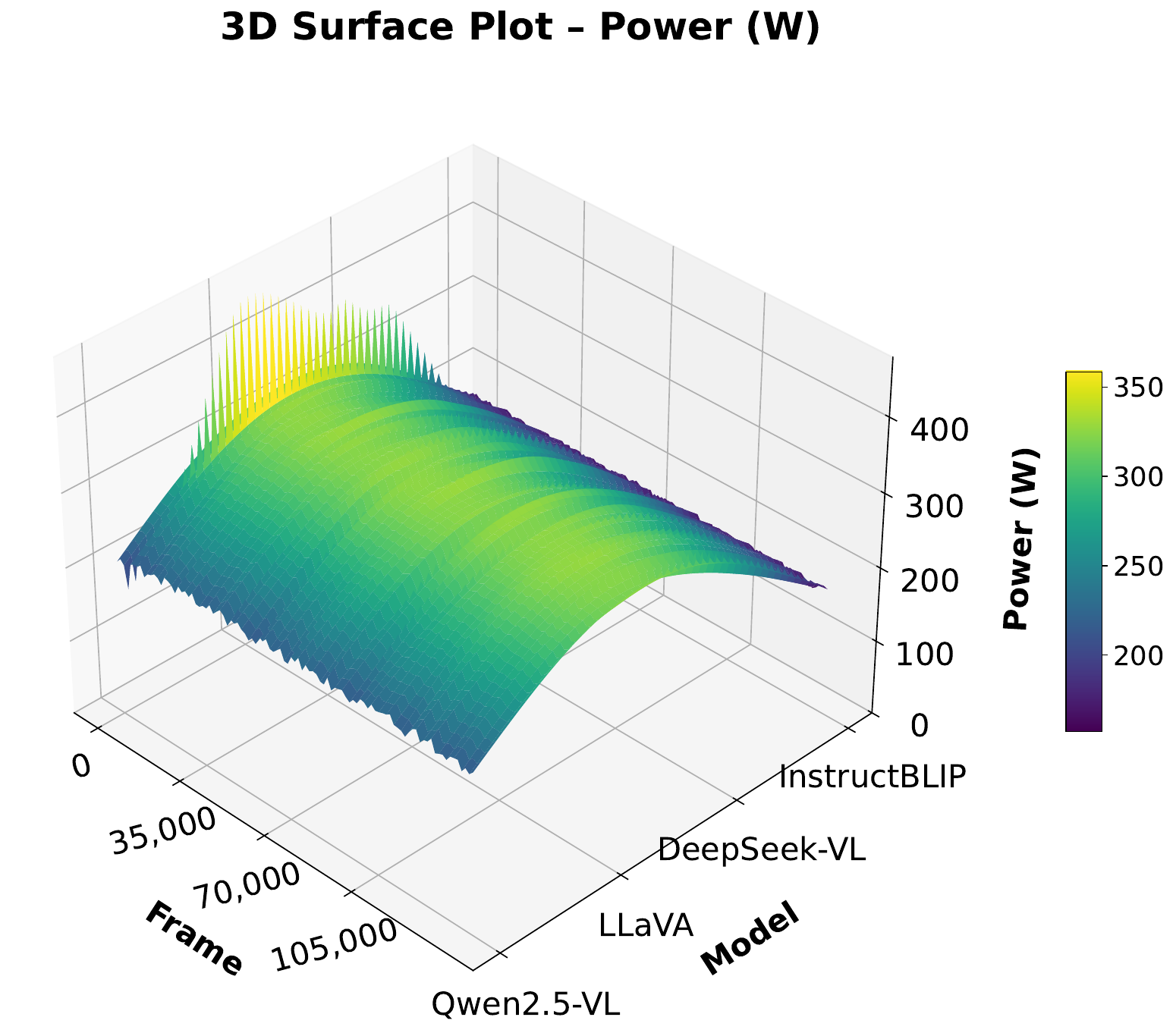}\label{fig:3d_power}}\hfill
    \subfloat[Confidence Score]{\includegraphics[width=0.31\textwidth, trim={0 0 0 3cm},clip]{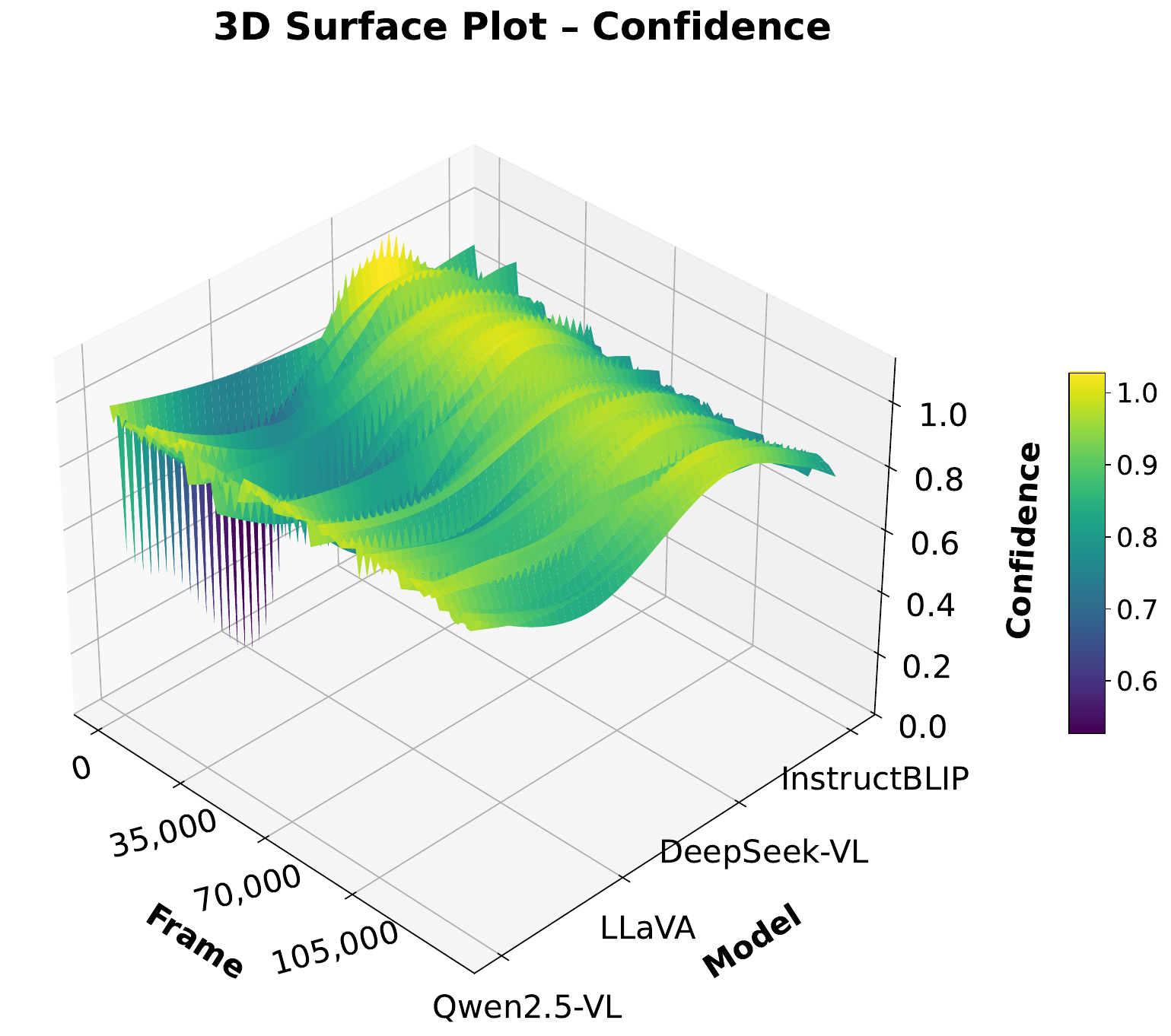}\label{fig:3d_conf}}\hfill
    \subfloat[Normalized Comparison]{\includegraphics[width=0.28\textwidth, trim={0 0 0 2cm},clip]{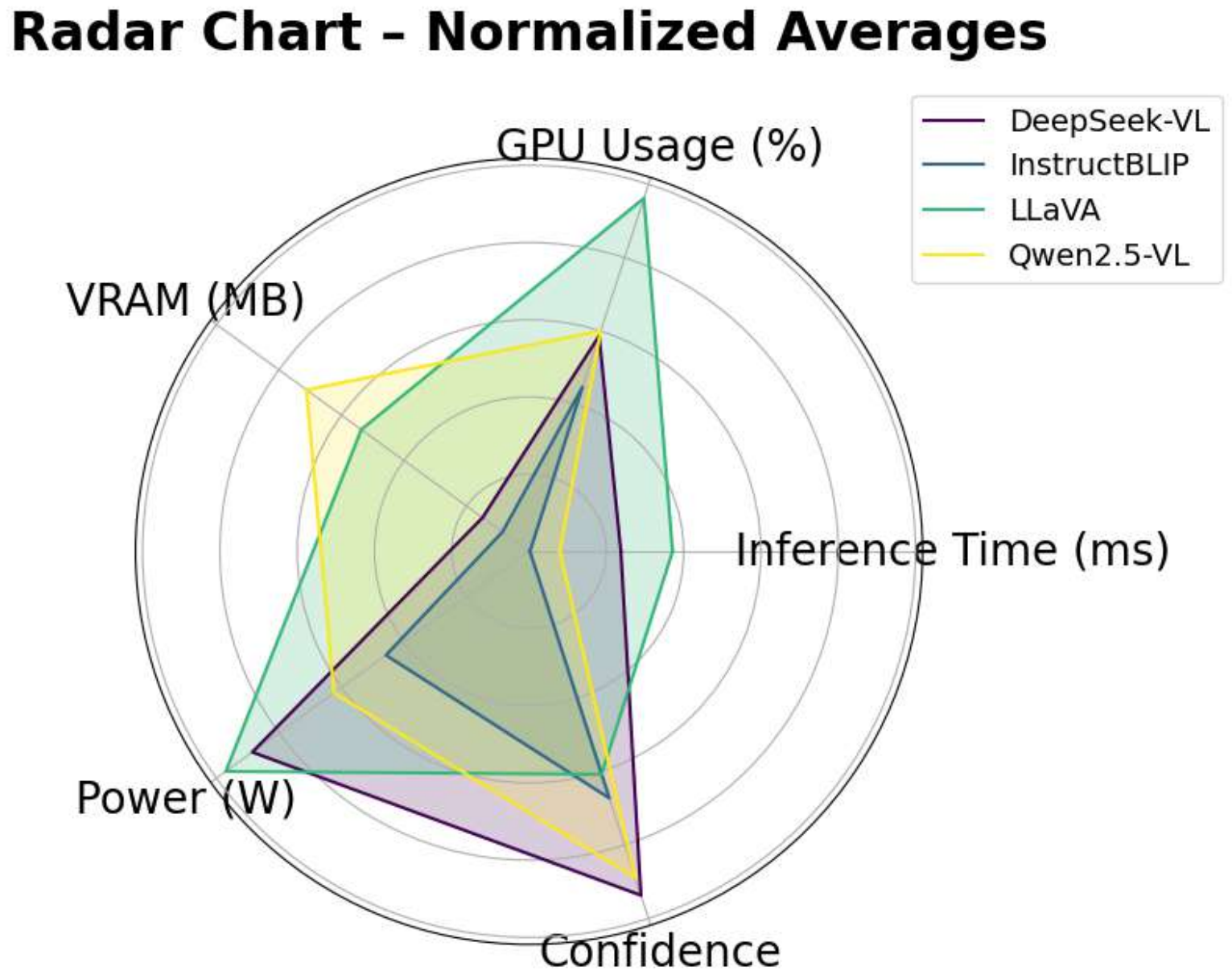}\label{fig:radar}}
    \caption{Computational profiles of the four models. Panels (a)--(e) report frame-level distributions across the full experimental dataset; panel (f) summarizes normalized model-level averages.}
    \label{fig:3d_overview}
\end{figure*}

{Figure~\ref{fig:3d_overview} reports the frame-level distributions of inference time, \ac{VRAM}, \ac{GPU} utilization, sampled power draw, and internal confidence. Figure~\ref{fig:radar} provides the corresponding normalized model-level comparison. These plots complement the aggregate statistics retained in the main text.}

\subsection{Category-level Inspection Time}
\label{app:time_by_category}
{Table~\ref{tab:time_savings_by_category} reports the complete category-level values underlying the averages in the main manuscript.}

\begin{table*}[!t]
\centering
\tiny
\caption{Processing time and percentage saved per category, averaged across prompt levels. The final row reports the arithmetic mean across categories.}
\label{tab:time_savings_by_category}
\begin{tabular}{l|ccccc|cccc}
\toprule
& \multicolumn{5}{c|}{Time in minutes} & \multicolumn{4}{c}{Perc. of time saved} \\
\textbf{Class} & Original & DeepSeek-VL & InstructBLIP & LLaVA & Qwen2.5-VL & DeepSeek-VL & InstructBLIP & LLaVA & Qwen2.5-VL \\
\midrule
Abuse & $107.50$ & $28.58$ & $9.71$ & $38.25$ & $17.25$ & $73.41$\% & $90.97$\% & $64.42$\% & $83.95$\% \\
Arrest & $165.22$ & $42.99$ & $14.97$ & $58.61$ & $24.13$ & $73.98$\% & $90.94$\% & $64.53$\% & $85.40$\% \\
Arson & $151.06$ & $39.22$ & $13.63$ & $53.50$ & $22.94$ & $74.04$\% & $90.98$\% & $64.58$\% & $84.81$\% \\
Assault & $72.19$ & $18.62$ & $6.49$ & $25.47$ & $10.69$ & $74.21$\% & $91.01$\% & $64.72$\% & $85.19$\% \\
Burglary & $261.78$ & $67.56$ & $23.66$ & $92.71$ & $38.31$ & $74.19$\% & $90.96$\% & $64.58$\% & $85.37$\% \\
Explosion & $140.22$ & $36.19$ & $12.67$ & $49.63$ & $20.32$ & $74.19$\% & $90.96$\% & $64.61$\% & $85.51$\% \\
Fighting & $143.83$ & $37.07$ & $13.02$ & $51.05$ & $20.70$ & $74.23$\% & $90.95$\% & $64.51$\% & $85.61$\% \\
RoadAccidents & $144.93$ & $37.06$ & $13.06$ & $51.04$ & $20.90$ & $74.43$\% & $90.99$\% & $64.78$\% & $85.58$\% \\
Robbery & $234.77$ & $60.59$ & $21.24$ & $83.34$ & $33.79$ & $74.19$\% & $90.95$\% & $64.50$\% & $85.61$\% \\
Shooting & $81.93$ & $21.11$ & $7.39$ & $28.89$ & $11.76$ & $74.23$\% & $90.98$\% & $64.74$\% & $85.65$\% \\
Shoplifting & $180.20$ & $46.63$ & $16.35$ & $63.85$ & $26.05$ & $74.12$\% & $90.93$\% & $64.57$\% & $85.54$\% \\
Stealing & $259.68$ & $67.06$ & $23.53$ & $91.92$ & $38.04$ & $74.18$\% & $90.94$\% & $64.60$\% & $85.35$\% \\
Vandalism & $81.76$ & $21.07$ & $7.42$ & $28.97$ & $11.75$ & $74.23$\% & $90.92$\% & $64.57$\% & $85.63$\% \\
\midrule
\textbf{Average} & \textbf{$155.77$} & \textbf{$40.29$} & \textbf{$14.09$} & \textbf{$55.17$} & \textbf{$22.82$} & \textbf{$74.13\%$} & \textbf{$90.96\%$} & \textbf{$64.59\%$} & \textbf{$85.32\%$} \\
\bottomrule
\end{tabular}
\end{table*}

\section{Notation and Symbols}
\label{sec:annex}
Table~\ref{tab:notation_summary} summarizes the notation used throughout the manuscript.
\begin{table*}[!t]
\caption{Summary of notation and symbols.}
\label{tab:notation_summary}
\centering
\scriptsize
\setlength{\tabcolsep}{3pt}
\renewcommand{\arraystretch}{1.05}
\begin{tabular}{|P{1.5cm}|p{15.6cm}|}
\hline
\multicolumn{2}{|c|}{\textbf{Dataset and Models}} \\
\hline
$\mathcal{V}$ & Set of videos = $\{\mathcal{V}_1, \dots, \mathcal{V}_{13}\}$, where each subset $\mathcal{V}_c$ contains all videos $v$ belonging to the $c$-th crime class. \\
\hline
$\mathcal{M}$ & Model space = \{\textit{Qwen2.5-VL}, \textit{InstructBLIP}, \textit{DeepSeek-VL}, and \textit{LLaVA-1.5}\} \\
\hline
$\mathcal{F}$ & Frame set, with frame $f \in \mathcal{F}$. \\
\hline
\multicolumn{2}{|c|}{\textbf{Inference and Efficiency}} \\
\hline
$T_{or}$ & Total original video duration (in minutes). \\
\hline
$T_{in}$ & Total inference time per model. \\
\hline
$T_s (\%)$ & $\left( \frac{T_{or} - T_{in}}{T_{or}} \right) \cdot 100$ = time saving percentage. \\
\hline
$r$ & Sampling rate, set to $1$ Hz in the experiments.\\
\hline
\multicolumn{2}{|c|}{\textbf{Model Output and Confidence}} \\
\hline
$\mathbf{s}_i$ & Logits output at token step $i$. \\
\hline
$\sigma(\mathbf{s}_i)$ & Softmax distribution at step $i$. \\
\hline
{$\mathcal{T}$} & {Total number of generated tokens in model output.} \\
\hline
{$\zeta$} & {Model confidence: $\frac{1}{\mathcal{T}} \sum_{i=1}^{\mathcal{T}} \max_j \sigma(\mathbf{s}_i)_j$---average of maximum softmax values across tokens.} \\
\hline
{$\tau$} & {Confidence threshold used to decide whether a positive response is included in the alert shortlist ($\zeta \geq \tau$).} \\
\hline
\multicolumn{2}{|c|}{\textbf{Prompt Readability (FK Grade)}} \\
\hline
$\text{FK Grade}$ & Flesch-Kincaid score of prompt complexity. \\
\hline
$a, b, c$ & Coefficients for FK: $a = 0.39$, $b = 11.8$, $c = -15.59$. \\
\hline
\multicolumn{2}{|c|}{\textbf{Hallucination Bucket Notation}} \\
\hline
{$B_k$} & {Number of evaluated configurations assigned to detection-rate bucket $k \in \{1,2,3,4,5\}$. A configuration is a video--prompt pair in the crime-video evaluation and a video--category--prompt triplet in the normal-video control.} \\
\hline
{$w_k$} & {Bucket weight: $w_1 = -0.5$, $w_2 = 0$, $w_3 = 0.5$, $w_4 = 1.0$, $w_5 = 1.5$.} \\
\hline
{$N$} & {Total number of evaluated configurations per model: $2{,}850$ for the crime-video evaluation and $1{,}950$ for the normal-video control.} \\
\hline
\multicolumn{2}{|c|}{\textbf{Detection and Scoring Metrics}} \\
\hline
{$C_{\mathscr{m}}$} & {Detection coverage: proportion of evaluated videos for which model $\mathscr{m} \in \mathcal{M}$ produces at least one positive response, averaged across the applicable categories and prompt levels.} \\
\hline
$D_{\mathscr{m}}$ & Per-frame detection rate: average ratio of frames marked positive by model ${\mathscr{m}}$. \\
\hline
$H_{\mathscr{m}}$ & Refined hallucination index: weighted average of detection bucket counts (see hallucination section). \\
\hline
$T_{\mathscr{m}}$ & Time saved; value normalized to [0,1]. Ratio of inference speed relative to original video duration.\\
\hline
$\text{CMUS}_{\mathscr{m}}$ & Composite Model Utility Score combining $C_{\mathscr{m}}$, $D_{\mathscr{m}}$, $T_{\mathscr{m}}$, and $H_{\mathscr{m}}$. \\
\hline
$\mathcal{P}$ & Prompt set = $\{p_1, p_2, p_3\}$ used to evaluate each frame at increasing levels of difficulty (easy, medium, hard). \\
\hline
$\mathcal{L}$ & Evaluation log storing the frame identifier, source timestamp, category, prompt, model, normalized response, confidence, and computational measurements for each inference step. \\
\hline
\multicolumn{2}{|c|}{\textbf{CMUS Weights}} \\
\hline
$\alpha, \beta$ & Reward weight for detection coverage (used as $\alpha = 1.0$), time savings (used as $\beta = 1.0$). \\
\hline
$\gamma, \delta$ & Signed penalty weights for detection rate ($\gamma = -0.5$) and refined hallucination index ($\delta = -1.0$). \\
\hline
\end{tabular}
\end{table*}

\clearpage
\balance
\bibliographystyle{IEEEtran}
\bibliography{references}